%% file: acl_latex.tex
\documentclass[11pt]{article}
\usepackage{marvosym}

\usepackage[preprint]{acl}

\usepackage{times}
\usepackage{latexsym}
\usepackage{minted}

\usepackage[T1]{fontenc}

\usepackage[utf8]{inputenc}

\usepackage{microtype}

\usepackage{inconsolata}

\usepackage{graphicx}

\usepackage{hyperref}       
\usepackage{url}            
\usepackage{booktabs}       
\usepackage{amsfonts}       
\usepackage{nicefrac}       
\usepackage{microtype}      
\usepackage{xcolor}         
\usepackage{amsmath}
\usepackage{amssymb}
\usepackage{nomencl}
\usepackage{glossaries}
\usepackage{xspace}
\usepackage{multirow}
\usepackage{pifont}
\usepackage{algorithm}
\usepackage{subfigure}
\usepackage{enumitem}
\usepackage{multicol}
\usepackage{caption}
\usepackage{array}
\usepackage{fp}
\usepackage{makecell}
\usepackage{flushend}
\usepackage{cases}
\usepackage{color}
\usepackage{algpseudocode}
\usepackage{listings}
\usepackage{mathrsfs}
\usepackage{longtable}
\usepackage{colortbl}
\usepackage{bm}
\usepackage{tabularray}
\usepackage{CJKutf8}

\newcommand{\ie}{\emph{i.e.,}\xspace}
\newcommand{\eg}{\emph{e.g.,}\xspace}

\newcommand{\ignore}[1]{}
\usepackage{tcolorbox}
\newtcolorbox{promptbox}[2][Prompt]{
colback=black!5!white,
arc=5pt, 
boxrule=0.5pt,
fonttitle=\bfseries,
title=#1, 
before upper={\scriptsize}, fontupper=\fontfamily{ptm}\selectfont,
colframe=#2, 
}

\lstdefinestyle{leanstyle}{
    basicstyle=\ttfamily\small,  
    columns=fullflexible,
    keepspaces=true,             
    breaklines=true,             
    extendedchars=true,
    literate=
        {ℕ}{{\ensuremath{\mathbb{N}}}}1
        {ℝ}{{\ensuremath{\mathbb{R}}}}1
        {∑}{{\ensuremath{\sum}}}1
        {≥}{{\ensuremath{\ge}}}1
        {≤}{{\ensuremath{\le}}}1
        {→}{{\ensuremath{\to}}}1
        {^2}{{$^2$}}1
        {:=}{{: \! =}}2,        
    frame=single,                
    backgroundcolor=\color{gray!5},
}

\title{Challenging the Boundaries of Reasoning:\\An Olympiad-Level Math Benchmark for Large Language Models}

\author{
Haoxiang Sun\textsuperscript{\rm{1}}, 
Yingqian Min\textsuperscript{\rm{2}},
Zhipeng Chen\textsuperscript{\rm{2}},\\
\textbf{Wayne Xin Zhao}\textsuperscript{\rm{2 \Letter}},
\textbf{Ji-Rong Wen}\textsuperscript{\rm{2}}\\
\textsuperscript{1} School of Information, Renmin University of China \\
\textsuperscript{2} Gaoling School of Artificial Intelligence, Renmin University of China \\
\texttt{\{hxiang.sun, batmanfly\}@gmail.com} \\
}

\begin{document}
\maketitle
\begingroup
\renewcommand\thefootnote{\Letter}
\footnotetext{Corresponding author.}
\endgroup
\begin{abstract}


The rapid advancement of large reasoning models has saturated existing math benchmarks, underscoring the urgent need for more challenging evaluation frameworks. {To address this, we introduce \textbf{OlymMATH}, a rigorously curated, Olympiad-level math benchmark comprising 350 problems, each with parallel English and Chinese versions. OlymMATH is the first benchmark to unify dual evaluation paradigms within a single suite: (1) \emph{natural language evaluation} through \textbf{OlymMATH-EASY} and \textbf{OlymMATH-HARD}, comprising 200 computational problems with numerical answers for objective rule-based assessment, and (2) \emph{formal verification} through \textbf{OlymMATH-LEAN}, offering 150 problems formalized in Lean 4 for rigorous process-level evaluation.}
All problems are manually sourced from printed publications to minimize data contamination, verified by experts, {and span four core domains}. Extensive experiments reveal the benchmark's significant challenge, and our analysis also uncovers consistent performance gaps between languages and identifies cases where models employ heuristic ``guessing'' rather than rigorous reasoning. To further support community research, we release 582k+ reasoning trajectories, a visualization tool, {and expert solutions at \href{ https://github.com/RUCAIBox/OlymMATH}{\texttt{https://github.com/RUCAIBox/OlymMATH}}.}

\end{abstract}

\input{introduction}

\input{dataset}
\input{experiments-nl}
\input{experiments-fl}
\input{conclusion}

\bibliography{custom}


\appendix

\input{appendix}

\end{document}

%% file: introduction.tex
\section{{Introduction}}

{The advent of large language models (LLMs)~\cite{zhao2025surveylargelanguagemodels} has marked a significant leap forward in the capabilities of artificial intelligence, with mathematical reasoning emerging as a pivotal and demanding area of research~\cite{deepseek_r1,openai2024openaio1card, chen-2025-still3}. Recently, the evaluation and enhancement of mathematical reasoning abilities have become a central focus in the development of LLMs~\cite{yang2024qwen25math}.}

Effective assessment of LLM reasoning necessitates \emph{reliable} and \emph{verifiable} evaluation benchmarks. {Reliability requires accurately designed problems with unambiguous solutions and minimized data contamination risk, ensuring trustworthy evaluation. For verifiability, two paradigms have emerged: (1) \emph{numerical-answer benchmarks} using rule-based verification (\eg \texttt{sympy}), which offer scalability but cannot assess reasoning quality; and (2) \emph{formal proof benchmarks} using theorem provers (\eg Lean, Isabelle), which provide rigorous process-level verification but require specialized formalization. An ideal benchmark suite should leverage both paradigms for comprehensive evaluation.}

{Existing benchmarks in both paradigms leave certain dimensions only partially addressed, and these gaps are not straightforwardly closed by recombining available resources. Among numerical-answer benchmarks, Olympiad-difficulty collections may not yet provide sufficient scale for robust statistical conclusions or enough headroom for the strongest current models (see Table~\ref{comp}); LLM-as-judge evaluation can be susceptible to evaluator hallucination, and reference solutions are occasionally incomplete. Among formal proof benchmarks, available datasets are English-only and drawn from well-known competitions with substantial online presence, warranting attention to potential contamination. More broadly, most existing benchmarks center on English, leaving multilingual reasoning comparatively less explored. Jointly ensuring consistent quality and low contamination across all these dimensions remains an open problem, as assembling web-crawled sources from different origins risks reintroducing the issues that motivate this work.}

To bridge the gap, we present \textbf{OlymMATH}, a rigorously curated, bilingual (English and Chinese) benchmark for Olympiad-level reasoning, comprising 350 {unique} problems organized into three {non-overlapping} subsets: \textbf{OlymMATH-EASY} and \textbf{OlymMATH-HARD} contain 100 {computational} problems each, split into \emph{easy} and \emph{hard} levels with parallel bilingual versions, requiring precise numerical answers for reliable and rule-based \texttt{sympy} verification. Additionally, \textbf{OlymMATH-LEAN} provides {a separate set of 150 problems formalized in Lean 4, accompanied by bilingual natural language statements and solutions}, enabling rigorous evaluation of automated theorem proving capabilities. Unlike proof-based benchmarks that rely on unreliable LLM-as-a-judge evaluation, OlymMATH-LEAN leverages the Lean language for fully automated and mathematically rigorous formal verification. Meanwhile, to prevent data leakage, problems were manually sourced from printed publications and verified by experts. The benchmark covers four major mathematical fields and adheres to the MATH or miniF2F dataset format for compatibility (see Figure~\ref{fig:splash}). 

\begin{figure}[t]
\noindent
\small

\begin{flushleft}

\begin{center}
\vspace{0.8em}
\large{OlymMATH-EASY / HARD}
\end{center}

\vspace{0.3em}

\textbf{Problem-EN:}
Find the remainder of $\sum_{k=0}^{1234}\binom{2016\times 1234}{2016k}$ modulo $2017^2$ (provide the value in the range $[0, 2017^2)$).

\vspace{0.1em}

\textbf{Answer:}\quad $1581330$.\quad\textbf{Subject:}\quad Number Theory.

\end{flushleft}

\begin{flushleft}

\vspace{-0.2em}
\hrule
\begin{center}
\vspace{0.5em}
\large{OlymMATH-LEAN}
\end{center}

\vspace{0.2em}

\textbf{Subject:}\quad Number Theory.\quad\textbf{Formal Statement:} 

\vspace{-0.2em}

{
\begin{verbatim}
theorem to_prove
  (n : Nat) (p : Nat) (hp : p.Prime)
  (hdiv : p | 2^n + 1) : p % 8 != 7 := by sorry
\end{verbatim}
}
\end{flushleft}

\caption{{Examples from our OlymMATH dataset.}}
\label{fig:splash}

\end{figure}

By leveraging OlymMATH, we conduct extensive experiments to evaluate the performance of state-of-the-art models. The results underscore our benchmark's difficulty, with advanced models like DeepSeek-R1, o3-mini, and Gemini 2.5 Pro Exp achieving only 19.5\%, 31.2\%, and 58.4\% accuracy, respectively, on OlymMATH-HARD (EN), indicating Olympiad-level math remains a significant challenge necessitating further research.
Our bilingual comparison showed a consistent performance gap, with higher accuracy on English problems versus Chinese, highlighting the need for multilingual evaluation.
Furthermore, case studies revealed models sometimes use heuristic ``guessing'' to reach answers without rigorous proofs. This underscores the importance of process-level inspection for accurate LLM capability assessment.

In summary, our contributions are as follows.

$\bullet$ {We introduce \textbf{OlymMATH}, the first Olympiad-level mathematical benchmark that unifies natural language problems and formal theorem proving within a single bilingual suite. OlymMATH comprises 350 unique problems, each available in both English and Chinese: \textbf{OlymMATH-EASY} and \textbf{OlymMATH-HARD} provide 200 computational problems with \texttt{sympy}-verifiable numerical answers, while \textbf{OlymMATH-LEAN} offers 150 problems formalized in Lean 4 for process-level verification—bridging the gap between outcome-based and reasoning-based evaluation.}

$\bullet$ Extensive experiments validate OlymMATH's reliability and strong discriminative power, while revealing critical model limitations including EN-ZH performance gaps and heuristic ``guessing'' that bypasses rigorous reasoning.

$\bullet$ We open-source 582,400 reasoning trajectories from 28 models, a visualization tool, and expert solutions to facilitate community research.

\section{{Related Work}}

\paragraph{Natural Language Math Benchmarks.}
The first paradigm of math reasoning evaluation relies on numerical-answer benchmarks with rule-based verification due to their simplicity and scalability. Early benchmarks such as GSM8K~\cite{cobbe2021training} and MATH~\cite{MATH} have been pivotal in advancing LLM reasoning capabilities~\cite{fang2024mathodysseybenchmarkingmathematicalproblemsolving, arora2023llmsadvancedenoughchallenging}, but are now largely saturated by slow-thinking models enhanced by long chain-of-thought fine-tuning~\cite{min-2024-still2} or reinforcement learning scaling~\cite{deepseek_r1}.

More challenging benchmarks face different limitations. The AIME dataset offers increased difficulty but suffers from small scale (\eg 30 problems from AIME 2025), compromising statistical reliability—a single problem shift can change accuracy by 3.33\%~\cite{hochlehnert2025soberlookprogresslanguage}, with binomial standard errors approximately 2.6$\times$ larger than those from a 200-problem benchmark. Moreover, rapidly improving models are approaching its measurement ceiling (\eg Gemini 2.5 Pro achieving 92\% \texttt{Pass@1} on AIME 2024), and its English-only focus neglects multilingual evaluation. OlympiadBench~\cite{OlympiadBench} provides a larger collection of problems, but its overall difficulty remains limited—even a 1.5B model (DeepScaleR-Preview~\cite{deepscaler2025}) achieves 50.0\% accuracy, indicating insufficient challenge for evaluating state-of-the-art reasoning models. Omni-MATH~\cite{gao2024omni} increases problem count via web crawling from AoPS, elevating data leakage risks, while its prevalent proof-based problems require LLM-as-judge evaluation rather than rule-based verification, reducing dependability. PolyMath~\cite{wang2025polymathevaluatingmathematicalreasoning} directly sources from widely publicized competitions (\eg AIME, CNMO, IMO) and existing datasets (\eg MGSM~\cite{shi2022language}, P-MMEval~\cite{zhang2024pmmevalparallelmultilingualmultitask}, HLE~\cite{phan2025humanitysexam}), risking high data leakage. AMO Bench~\cite{an2025amobenchlargelanguagemodels} targets advanced Olympiad problems but has limited scale.

In contrast, OlymMATH provides 200 challenging Olympiad-level problems with lower contamination risk through manual curation from printed publications, larger scale for statistical reliability, and bilingual versions for thorough evaluation.

\paragraph{Formal Language Math Benchmarks.}
Beyond assessing final answers, understanding \emph{how} models arrive at solutions—distinguishing rigorous derivation from heuristic shortcuts—is equally important. Formal theorem proving benchmarks address this by requiring machine-checkable proofs. miniF2F~\cite{zheng2022minif2fcrosssystembenchmarkformal} provides Olympiad-level problems formalized in multiple proof assistants, while ProofNet~\cite{azerbayev2023proofnetautoformalizingformallyproving} focuses on undergraduate mathematics. FIMO~\cite{liu2023fimochallengeformaldataset} and PutnamBench~\cite{tsoukalas2024putnambenchevaluatingneuraltheoremprovers} target competition mathematics at different levels. However, existing formal benchmarks typically source from well-known competitions (\eg IMO shortlist for FIMO, Putnam for PutnamBench) with high online exposure, facing similar contamination risks. Moreover, all existing formal benchmarks are English-only. OlymMATH-LEAN addresses these gaps with bilingual Lean 4 formalizations sourced from printed publications.


%% file: dataset.tex
\section{Benchmark Construction}

\begin{table}[t]
\centering
\small
\textbf{(a) Natural Language Benchmarks}
\vspace{0.3em}
\begin{tblr}{
  colspec = {l c c c},
  row{1} = {m},
  colsep = 3.5pt,
  hline{1,11} = {-}{0.08em},
  hline{2,9} = {-}{0.05em},
}
\textbf{Name}           & {\textbf{\# Prob.} \\ \textbf{(\# Lang.)}}& \textbf{Eval.}     & \textbf{Difficulty}   \\
AIME 24,25          & 30 (1)   & Rule        & Olympiad    \\
HMMT          & 30 (1)        & Rule        & Olympiad    \\
USAMO 2025 & 6 (1)& LLM & Olympiad \\
OlympiadBench & 2133 (2)          & Rule & Olympiad \\
Omni-MATH     & 4428 (1)             & LLM  & Olympiad \\
PolyMath     & 500 (18)            & Rule & Olympiad\\
AMO Bench    & 50 (1)        & Rule \& LLM & Olympiad\\
\textbf{EASY (Ours)}    & 100 (2)    & Rule & Olympiad\\
\textbf{HARD (Ours)}    & 100 (2)   & Rule & Olympiad\\
\end{tblr}

\vspace{1em}

\textbf{(b) Formal Language Benchmarks}
\vspace{0.3em}

\begin{tblr}{
  colspec = {l c c c},
  colsep = 3.5pt,
  hline{1,7} = {-}{0.08em},
  hline{2,6} = {-}{0.05em},
}
\textbf{Name}           & \textbf{\# Prob.} & \textbf{Lang.}   & \textbf{Difficulty}   \\
miniF2F     & 488         &   EN       & Olympiad\\
ProofNet     & 371         &   EN     & Undergrad\\
FIMO     & 149         &   EN      & Olympiad\\
PutnamBench      & 640       &   EN      & Undergrad Comp.\\
\textbf{LEAN (Ours)}   & 150        & EN \& ZH   & Olympiad
\end{tblr}
\caption{Comparison of existing benchmarks. EN and ZH denote English and Chinese, respectively.}
\label{comp}
\end{table}

{In this section, we describe OlymMATH in detail, including its construction methodology, problem composition, categorical distribution, and evaluation approach. Table~\ref{comp} presents a comparison with existing mathematical reasoning benchmarks. Existing benchmarks typically focus on either \emph{natural language problems} with numerical answers or \emph{formal theorem proving}, but not both. OlymMATH bridges this gap by being the first Olympiad-level benchmark to integrate both paradigms within a unified bilingual framework: OlymMATH-EASY and OlymMATH-HARD provide 200 natural language problems requiring precise numerical answers for scalable rule-based verification, while OlymMATH-LEAN offers 150 problems with Lean 4 formalizations enabling rigorous process-level verification. This dual-paradigm design allows comprehensive assessment of both outcome correctness and reasoning rigor, addressing the limitations of relying on either paradigm alone.}



\subsection{Contamination Analysis \& Verification}

\paragraph{Contamination Analysis} {Constructing a reliable benchmark requires mitigating data contamination. OlymMATH comprises 350 problems curated from printed resources (specialized magazines and textbooks), intentionally excluding online repositories to minimize prior digital exposure, unlike existing benchmarks drawing from well-known competitions (\eg FIMO using IMO shortlist, PutnamBench using Putnam, Omni-MATH using AoPS).}

{For quantitative leakage analysis, we followed Omni-MATH, using $n$-gram accuracy metric~\cite{xu2024benchmarkingbenchmarkleakagelarge}: for each sample, the problem and answer are concatenated; 5 starting points are uniformly sampled; and the model's ability to predict the subsequent 5-gram is evaluated. Leakage risk is quantified by comparing $n$-gram accuracy on the original dataset against 3 LLM-rewritten versions (Gemini 2.5 Flash Preview Thinking~\cite{geminiflash}), with the normalized difference $\delta$ indicating model familiarity with original versus rewritten data. Since $\delta$'s absolute value depends on the rewriting LLM, assessing leakage risk requires \emph{relative comparison} of $\delta$ between benchmarks. Results in Table~\ref{leak} show lower contamination risk for OlymMATH than PolyMath, establishing OlymMATH as a more reliable benchmark for evaluating LLMs' true mathematical capabilities.}

\begin{table}[t]
\centering
\small

\resizebox{\linewidth}{!}{
\begin{tblr}{
  colspec = {l c c c},
  hline{1,6} = {-}{0.08em},
  hline{2} = {-}{0.05em},
  hline{4} = {-}{0.03em, dashed},
}
\textbf{Model (Base)} & \textbf{Lang.} & \textbf{PolyMath} & \textbf{OlymMATH} \\
\SetCell[r=2]{l,m} {InternLM2-Math-7B \\ \cite{ying2024internlmmath}} 
      & EN & 34.84\% & \textbf{0.90\%} \\
      & ZH & 12.29\% & \textbf{0.88\%} \\
\SetCell[r=2]{l,m} {Qwen2.5-7B \\ \cite{qwen2.5}} 
      & EN & 38.81\% & \textbf{17.59\%} \\
      & ZH & 10.27\% & \textbf{3.42\%} \\
\end{tblr}
}
\caption{Results of leakage analysis. The lower value is bolded. OlymMATH exhibits lower $\delta$ values than PolyMath per language, indicating a lower leakage risk.}
\label{leak}
\end{table}

\paragraph{Verification} To enhance dataset reliability, we invited a China Mathematical Olympiad silver medalist and two provincial first-prize winners to verify and revise the problems and solutions in OlymMATH-EASY and HARD. Since the answers to the problems were already provided, the verification difficulty was reduced, making the expertise of reviewers sufficient for this task. Each problem was reviewed by at least two reviewers. Additionally, we have published official solutions for challenging problems for community oversight.

{For OlymMATH-LEAN, we leverage the Lean server for automatic verification. Raw problems and solutions are first cleaned by Claude Opus 4.5~\cite{claude4.5opus} for \LaTeX{} format correction, then undergo three independent verification rounds using DeepSeek V3.2 Speciale~\cite{deepseekai2025deepseekv32pushingfrontieropen} to check translation accuracy, statement precision, and solution rigor. A Claude Opus 4.5-based agent (see Appendix~\ref{section:prompts} for details) then iteratively interacts with a Kimina Lean REPL server~\cite{santos2025kiminaleanserverhighperformance} in an isolated sandbox, refining code based on compiler feedback until successful compilation. Compiled formalizations are validated by three independent Gemini 3.0 Flash~\cite{gemini3.0flash} calls for semantic alignment, and formalizations containing \texttt{axiom} declarations receive additional human expert reviews.}

\subsection{Problem Categories and Distribution}

{OlymMATH problems span four key high-school Olympiad mathematical fields—algebra, geometry, number theory, and combinatorics—classified by human experts for reliability. Problems are selected for their challenge, suitability for simple-answer verification, and topic diversity (\eg inequalities, sequences, and more in algebra). Figure-based problems within this set are text-reformulated for LLM compatibility, with non-convertible ones excluded (\eg Figure \hyperref[tab:geo-prob]{5} in Appendix).}

For refined evaluation, computational problems are categorized by difficulty: \emph{easy}, designed to challenge standard prompting in mainstream models, and \emph{hard}, tailored to test advanced reasoning (\eg slow-thinking modes) in state-of-the-art models. {Additionally, OlymMATH-LEAN provides 150 problems with Lean 4 (Mathlib v4.24.0) formalizations for process-level verification.} The distribution details are described in Table~\ref{datatype}.

\begin{table}[t]
\centering
\small

\SetTblrInner{rowsep=2pt}
\begin{tblr}{
  colspec = {l *{3}{c}},
  column{1} = {l, font=\linespread{0.9}\selectfont},
  rows = {m},
  cell{1}{1} = {r=2}{m},
  cell{1}{2} = {c=3}{c},
  hline{1,8} = {-}{0.08em},
  hline{3,7} = {-}{0.05em},
  hline{2} = {2-4}{0.03em},
}
\textbf{Category} & \textbf{\# Problems} &  &   \\
         & \textbf{\textsc{Easy}} & \textbf{\textsc{Hard}} & \textbf{\textsc{Lean}}  \\
{Algebra (Alg.)\\ {\scriptsize\itshape Inequality, Trigonometry, etc.}}       
         & 25 & 25 & 79 \\
{Geometry (Geo.)\\ {\scriptsize\itshape Solid \& Analytic Geometry, etc.}}      
         & 33 & 25 & 15 \\
{Number Theory (Num.)\\ {\scriptsize\itshape Diophantine Equation, etc.}} 
         & 13 & 25 & 42 \\
{Combinatorics (Com.)\\ {\scriptsize\itshape Graph Theory, Permutation, etc.}}          
         & 29 & 25 & 14  \\
Total    & 100 & 100 & 150 \\
\end{tblr}
\caption{The distribution of OlymMATH by category.}
\label{datatype}
\end{table}

\subsection{Format and Verification Methodology}

OlymMATH adopts MATH and miniF2F dataset format (see Figure \ref{fig:splash}) for seamless integration with existing pipelines and enhancing clarity and processing efficiency. All problems are text-based, including geometry problems reformulated from diagrams to align with LLM evaluation, as mentioned previously.
{For consistent, objective assessment, answers to computational problems are restricted to real numbers and intervals, \eg ``$[\sqrt{33}, +\infty)$'', while excluding ambiguous formats such as set operations, variables, complex numbers, and texts (see Table~\ref{format} in Appendix for details). This design enables reliable \texttt{sympy}-based and formal Lean server verification.}


To make the evaluation more challenging, OlymMATH includes problems with multiple numerical answers. These problems are modified to require a summary of all potential outcomes (\eg sums, sums of squares; see Figure \hyperref[tab:case3]{6} in Appendix). This method effectively assesses whether models can consider all possible answers, thereby providing a robust evaluation of their reasoning capabilities.

\subsection{Bilingual Extension}

{Originating from Chinese-language problems, OlymMATH provides both original Chinese and translated English versions for bilingual evaluation. Our translation pipeline employs Claude Sonnet 3.7~\cite{claude3.7} for initial translation, iterative refinement with GPT-4o~\cite{openai2024gpt4ocard}, and human verification by two expert annotators to ensure mathematical accuracy and linguistic fluency. OlymMATH-LEAN similarly provides bilingual natural language statements alongside Lean formalizations, supporting research in multilingual reasoning and informal-formal translation.}

%% file: experiments-nl.tex
\section{Experiments}

In this section, we assess the performance of leading reasoning models using OlymMATH and provide a detailed analysis of their capabilities.

\subsection{{Natural Language: \textsc{Easy} \& \textsc{Hard} Subset}}

{We first evaluate models on the natural language subsets, where problems require numerical answers verified via rule-based matching.}

\subsubsection{Experimental Setup}

\paragraph{Models.} 
We assess representative LLMs for a thorough evaluation. For open-source models, we investigated recent work on reasoning models, and evaluated DeepSeek-R1~\cite{deepseek_r1}, STILL-3-Preview~\cite{Slow_Thinking_with_LLMs_3_Preview}, DeepScaleR-Preview~\cite{deepscaler2025}, QwQ~\cite{qwq32b}, Light-R1~\cite{wen2025lightr1curriculumsftdpo}, OpenThinker2~\cite{openthoughts}, Skywork-OR1~\cite{skywork-or1-2025}, GLM-Z1-Air~\cite{glm2024chatglm}, AceMath-RL~\cite{acemath2024}, OpenMath-Nemotron~\cite{moshkov2025aimo2winningsolutionbuilding}, and Qwen3~\cite{qwen3}. For closed-source models, we evaluate o3-mini~(high)~\cite{openai2025o3mini} and Gemini 2.5 Pro Exp 0325~\cite{geminiproexp}.

\definecolor{consbg}{gray}{0.95}
\begin{table*}[t]
\centering
\setlength{\tabcolsep}{1.5pt}

\resizebox{\linewidth}{!}{
\begin{tabular}{l c >{\columncolor{consbg}}c c >{\columncolor{consbg}}c c >{\columncolor{consbg}}c c >{\columncolor{consbg}}c c >{\columncolor{consbg}}c | c >{\columncolor{consbg}}c c >{\columncolor{consbg}}c c >{\columncolor{consbg}}c c >{\columncolor{consbg}}c c >{\columncolor{consbg}}c }
\toprule
\multirow{3.5}{*}{\textbf{Model}} & \multicolumn{10}{c|}{\textbf{OlymMATH-HARD}} & \multicolumn{10}{c}{\textbf{OlymMATH-EASY}} \\
\cmidrule(lr){2-11} \cmidrule(lr){12-21}
& \multicolumn{2}{c}{\textbf{Alg.}} & \multicolumn{2}{c}{\textbf{Geo.}} & \multicolumn{2}{c}{\textbf{Num.}} & \multicolumn{2}{c}{\textbf{Com.}} & \multicolumn{2}{c|}{\textbf{Avg.}} & \multicolumn{2}{c}{\textbf{Alg.}} & \multicolumn{2}{c}{\textbf{Geo.}} & \multicolumn{2}{c}{\textbf{Num.}} & \multicolumn{2}{c}{\textbf{Com.}} & \multicolumn{2}{c}{\textbf{Avg.}} \\
& \texttt{P@1} & \texttt{C@k} & \texttt{P@1} & \texttt{C@k} & \texttt{P@1} & \texttt{C@k} & \texttt{P@1} & \texttt{C@k} & \texttt{P@1} & \texttt{C@k} & \texttt{P@1} & \texttt{C@k} & \texttt{P@1} & \texttt{C@k} & \texttt{P@1} & \texttt{C@k} & \texttt{P@1} & \texttt{C@k} & \texttt{P@1} & \texttt{C@k} \\
\midrule
\multicolumn{21}{c}{\textit{\textbf{English (EN)}}} \\
\midrule
Qwen3 (0.6B, Think) & 2.5 & 0.0 & 2.1 & 4.0 & 6.6 & 8.0 & 0.2 & 0.0 & {2.8} & {3.0} & 15.5 & 20.0 & 5.6 & 15.2 & 24.5 & 38.5 & 5.2 & 6.9 & {10.4} & {17.0} \\
DS-R1-Distill (1.5B) & 1.9 & 0.0 & 1.8 & 0.0 & 1.8 & 0.0 & 0.4 & 0.0 & 1.5 & 0.0 & 20.8 & 40.0 & 12.6 & 21.2 & 32.6 & 61.5 & 8.2 & 24.1 & 16.0 & 32.0 \\
Qwen3 (4B, Think) & 18.1 & 20.0 & 14.8 & 12.0 & 19.8 & 28.0 & 3.1 & 4.0 & {13.9} & {16.0} & 76.4 & 92.0 & 79.1 & 97.0 & 85.1 & 84.6 & 57.1 & 72.4 & {72.8} & {87.0} \\
DS-R1-Distill (7B) & 15.6 & 36.0 & 12.6 & 24.0 & 13.1 & 24.0 & 3.1 & 4.0 & 11.1 & 22.0 & 52.8 & 84.0 & 49.6 & 84.8 & 62.5 & 84.6 & 33.9 & 58.6 & 47.5 & 77.0 \\
Qwen3 (30B-A3B, Think) & 38.8 & 44.0 & 33.8 & 44.0 & 26.7 & 36.0 & 5.9 & 4.0 & {26.3} & {32.0} & 91.4 & 100.0 & 92.9 & 100.0 & 90.9 & 92.3 & 75.6 & 93.1 & {87.2} & {97.0} \\
DS-R1-Distill (32B) & 22.4 & 32.0 & 21.4 & 24.0 & 20.3 & 40.0 & 3.4 & 4.0 & 16.9 & 25.0 & 73.6 & 100.0 & 71.8 & 97.0 & 84.5 & 92.3 & 49.0 & 69.0 & 67.3 & 89.0 \\
QwQ (32B) & 32.9 & 28.0 & 26.6 & 36.0 & 26.7 & 44.0 & 6.2 & 4.0 & {23.1} & 28.0 & 91.8 & 100.0 & 87.0 & 100.0 & 95.0 & 100.0 & 69.0 & 89.7 & {84.0} & {97.0} \\
\textcolor{gray}{Qwen3 (235B-A22B, Think)} & 48.0 & 52.0 & 49.5 & 60.0 & 38.0 & 36.0 & 10.5 & 16.0 & {36.5} & {41.0} & 93.5 & 100.0 & 92.4 & 100.0 & 99.0 & 100.0 & 81.9 & 93.1 & {90.5} & {98.0} \\
\textcolor{gray}{DeepSeek R1} & 30.0 & 40.0 & 25.5 & 32.0 & 18.5 & 24.0 & 4.0 & 4.0 & 19.5 & 25.0 & 90.5 & 100.0 & 82.2 & 97.0 & 94.2 & 100.0 & 60.8 & 72.4 & 79.6 & 91.0 \\
\textcolor{gray}{OpenAI o3-mini (high)} & 29.5 & 32.0 & 29.0 & 44.0 & 49.5 & 60.0 & 17.0 & 20.0 & {31.2} & {39.0} & 93.0 & 92.0 & 89.8 & 100.0 & 97.1 & 100.0 & 89.2 & 96.6 & {91.4} & \textbf{97.0} \\
\textcolor{gray}{Gemini 2.5 Pro Exp 0325} & 71.5 & 76.0 & 75.5 & 84.0 & 59.0 & 72.0 & 27.5 & 36.0 & {\textbf{58.4}} & {\textbf{67.0}} & 92.0 & 100.0 & 97.0 & 100.0 & 98.1 & 100.0 & 84.5 & 89.7 &{\textbf{92.2}} & \textbf{{97.0}} \\
\midrule
\multicolumn{21}{c}{\textit{\textbf{Chinese (ZH)}}} \\
\midrule
Qwen3 (0.6B, Think) & 2.6 & 4.0 & 0.8 & 0.0 & 4.4 & 4.0 & 0.0 & 0.0 & {1.9} & {2.0} & 9.9 & 8.0 & 2.8 & 3.0 & 12.0 & 15.4 & 1.3 & 3.4 & {5.4} & {6.0} \\
DS-R1-Distill (1.5B) & 1.8 & 0.0 & 1.3 & 0.0 & 1.1 & 0.0 & 0.0 & 0.0 & 1.0 & 0.0 & 13.7 & 20.0 & 6.3 & 9.1 & 20.9 & 30.8 & 2.6 & 0.0 & 9.0 & 12.0 \\
Qwen3 (4B, Think) & 12.5 & 20.0 & 7.0 & 8.0 & 12.6 & 24.0 & 0.9 & 0.0 & {8.3} & {13.0} & 70.8 & 88.0 & 61.0 & 75.8 & 74.8 & 92.3 & 41.8 & 51.7 & {59.7} & {74.0} \\
DS-R1-Distill (7B) & 6.1 & 8.0 & 7.9 & 12.0 & 6.6 & 8.0 & 0.6 & 0.0 & 5.3 & {7.0} & 38.0 & 64.0 & 30.8 & 51.5 & 49.2 & 61.5 & 18.7 & 27.6 & 31.5 & 49.0 \\
Qwen3 (30B-A3B, Think) & 35.6 & 40.0 & 24.1 & 28.0 & 18.1 & 24.0 & 2.7 & 4.0 & {20.1} & {24.0} & 87.8 & 92.0 & 84.7 & 97.0 & 91.3 & 100.0 & 61.9 & 65.5 & {79.7} & {87.0} \\
DS-R1-Distill (32B) & 6.5 & 0.0 & 5.4 & 4.0 & 10.6 & 12.0 & 0.7 & 0.0 & 5.8 & 4.0 & 45.2 & 52.0 & 41.8 & 63.6 & 60.2 & 69.2 & 26.0 & 37.9 & 40.4 & 54.0 \\
QwQ (32B) & 20.9 & 24.0 & 15.9 & 16.0 & 17.6 & 24.0 & 2.0 & 0.0 & 14.1 & 16.0 & 85.4 & 96.0 & 76.6 & 97.0 & 92.9 & 100.0 & 53.8 & 69.0 & {74.3} & {89.0} \\
\textcolor{gray}{Qwen3 (235B-A22B, Think)} & 36.5 & 48.0 & 43.5 & 48.0 & 28.5 & 32.0 & 4.0 & 8.0 & {28.1} & {34.0} & 91.0 & 100.0 & 90.2 & 97.0 & 94.2 & 100.0 & 78.4 & 89.7 & {87.5} & {96.0} \\
\textcolor{gray}{DeepSeek R1} & 20.0 & 24.0 & 25.0 & 28.0 & 17.0 & 16.0 & 1.5 & 0.0 & 15.9 & 17.0 & 79.5 & 96.0 & 74.6 & 84.8 & 88.5 & 92.3 & 49.6 & 55.2 & 70.4 & 80.0 \\
\textcolor{gray}{OpenAI o3-mini (high)} & 31.5 & 40.0 & 32.5 & 44.0 & 48.5 & 56.0 & 19.0 & 28.0 & {32.9} & {42.0} & 93.0 & 96.0 & 89.4 & 100.0 & 99.0 & 100.0 & 85.8 & 93.1 & {90.5} & \textbf{97.0} \\
\textcolor{gray}{Gemini 2.5 Pro Exp 0325} & 65.0 & 76.0 & 78.0 & 80.0 & 53.5 & 56.0 & 25.0 & 40.0 & {\textbf{55.4}} & {\textbf{63.0}} & 90.5 & 96.0 & 93.2 & 93.9 & 100.0 & 100.0 & 84.1 & 86.2 & {\textbf{90.8}} & {{93.0}} \\
\bottomrule
\end{tabular}
}
\caption{{Model performance on OlymMATH sorted by model size. The abbreviations ``Alg.'', ``Geo.'', etc. represent the four categories in OlymMATH. Models sampled only 8 times are marked in \textcolor{gray}{gray} to indicate potential instability. For brevity, only representative models are shown; see Table~\ref{performance-en-full} and Table~\ref{performance-zh-full} in Appendix for complete results.}}
\label{performance}
\end{table*}

\paragraph{Evaluation Details.}

Our evaluation pipeline generates 64 responses per problem for each model, except for resource-intensive models (\ie OpenMath-Nemotron-32B, Qwen3-235B-A22B, GLM-Z1-Air, DeepSeek-R1, o3-mini (high), and Gemini 2.5 Pro Exp), which are limited to 8 samples due to resource limitations and the relatively large scale of our dataset. For \texttt{Pass@1}, we compute mean accuracy across all sampled responses; for \texttt{Cons@64} and \texttt{Cons@8}, we apply majority voting to determine a consensus answer per problem. Following established practices~\cite{deepseek_r1,qwq32b}, locally-evaluated models use \texttt{temperature} = 0.6, \texttt{top\_p} = 0.95, \texttt{min\_p} = 0, and \texttt{max\_tokens} = 32768, while API-evaluated models (\ie GLM-Z1-Air, DeepSeek-R1, o3-mini (high), and Gemini 2.5 Pro Exp) use maximum available \texttt{max\_tokens} to fully leverage their reasoning capabilities. We open-source all 582,400 samples, an online visualization tool, and standard solutions for challenging problems to support community analysis of LLM reasoning patterns and characteristics (see Appendix~\ref{section:eval} for further information).

\subsubsection{{Evaluation Results}}

In this part, we present the evaluation results of OlymMATH (EN) and OlymMATH (ZH) in Table~\ref{performance}. {Due to space constraints, we include only representative models in the main text, and full results are provided in Tables~\ref{performance-en-full} and~\ref{performance-zh-full} in Appendix.}

\begin{figure}[t]
    \centering
    \includegraphics[width=\linewidth]{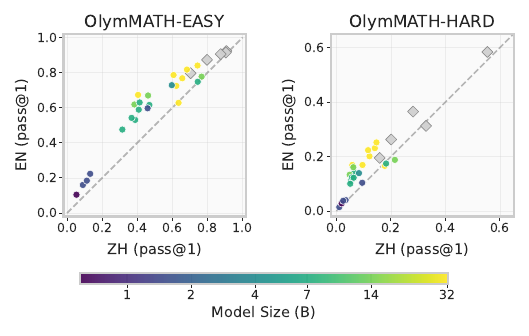}
    \caption*{{Figure 2: {\texttt{Pass@1} on OlymMATH EN (y) vs. ZH (x), the dashed line shows parity. Points above favor EN, below favor ZH. Solid circles (local dense models, colored by size) indicate larger models trend towards higher accuracy. Diamonds are MoE or closed-source models.}}}
    \label{lang}
\end{figure}

First, we observe that all tested models exhibit relatively poor performance, with even OpenAI o3-mini (high) and Gemini 2.5 Pro Exp achieving only 31.2\% and 58.4\% on OlymMATH-HARD (EN). This underscores the high overall difficulty of our benchmark, which demands stronger reasoning abilities and a deeper understanding of mathematical knowledge. In contrast, the performance of these advanced reasoning models on OlymMATH-EASY (EN) is more modest and comparable to that on AIME 2024, suggesting that OlymMATH-EASY is well-suited for evaluating the capabilities of less advanced reasoning models. 

Second, by comparing the performance of LLMs on bilingual versions of OlymMATH, we find that language can influence the reasoning performance of LLMs to some extent (see Figure \hyperref[lang]{2}). 
Overall, all models tend to achieve higher performance on the English benchmarks. A potential reason for this is that English corpora still dominate existing pre-training datasets, making the English-based task-solving capabilities of LLMs generally superior compared to other languages. {Prior work has documented cross-lingual reasoning gaps in word grouping~\citep{guerra-solano-etal-2025-think}, multilingual reasoning paths~\citep{tam2025languagemattersmultilingualinput}, and thinking trace languages~\citep{qi-etal-2025-models}. OlymMATH extends these findings to Olympiad-level mathematics: Wilcoxon signed-rank tests on our released 582k trajectories from 14 models (1.5B, 7B, 14B) confirm that the EN-ZH gap is statistically significant across all subjects and difficulty levels. Trajectory analysis further shows that extraction failures are disproportionately frequent among incorrect ZH responses, pointing to a language-specific presentation issue distinct from reasoning errors.}


{Third, to provide insights into model robustness beyond \texttt{Pass@1}, we report \texttt{Pass@k} for DeepSeek-R1-Distill-Qwen series in Table~\ref{tab-passk}. The results reveal substantial gains from increased sampling: 7B model improves from 11.1\% (\texttt{Pass@1}) to 74.0\% (\texttt{Pass@64}) on EN-HARD, indicating that correct solutions exist within the model's capability but require multiple attempts to surface. However, the gap between \texttt{Pass@64} and \texttt{Cons@64} (74.0\% vs. 22.0\%) suggests significant inconsistency—models can solve problems but often fail to do so reliably. These results support the use of \texttt{Pass@k} as a more comprehensive metric for probing the reasoning capability boundaries of LLMs~\cite{yue2025doesreinforcementlearningreally}.}

\begin{table}[t]
\centering
\small

\begin{tblr}{
  colspec = {l l c c c c c},
  hline{1,8} = {-}{0.08em},
  hline{2} = {-}{0.05em},
  hline{4,6} = {-}{0.03em, dashed},
}
\textbf{Model} & \textbf{Subset} & \textbf{\texttt{P@1}} & \textbf{\texttt{P@4}} & \textbf{\texttt{P@16}} & \textbf{\texttt{P@64}} & \textbf{\texttt{C@64}} \\
\SetCell[r=2]{l,m} 1.5B & \textsc{Easy} & 16.0 & 37.5 & 62.2 & 78.0 & 32.0 \\
      & \textsc{Hard} & 1.5 & 5.1 & 14.2 & 30.0 & 0.0 \\
\SetCell[r=2]{l,m} 7B & \textsc{Easy} & 47.5 & 78.4 & 91.8 & 97.0 & 77.0 \\
      & \textsc{Hard} & 11.1 & 29.6 & 53.4 & 74.0 & 22.0 \\
\SetCell[r=2]{l,m} 32B & \textsc{Easy} & 67.3 & 90.8 & 97.4 & 100.0 & 89.0 \\
      & \textsc{Hard} & 16.9 & 38.7 & 59.0 & 75.0 & 25.0 \\
\end{tblr}
\caption{\texttt{Pass@k} and \texttt{Cons@64} for DS-R1-Distill series on OlymMATH-EASY and HARD (in English).}
\label{tab-passk}
\end{table}

\subsubsection{{Benchmark Comparison}}

\begin{table}[t]
\centering
\small
\begin{tblr}{
  colspec = {l c c c c c},
  hline{1,11} = {-}{0.08em},
  hline{2} = {-}{0.05em},
  row{1} = {font=\bfseries},
  colsep = 2pt
}
Model & AIME & OBench & Omni & \textsc{Easy} & \textsc{Hard} \\
STILL-3-Pre. (1.5B) & 32.5 & 45.4 & - & 18.4 & 3.8 \\
DScaleR-Pre. (1.5B) & 43.1 & 50.0 & - & 22.3 & 4.1 \\
GPT-4o & 13.1 & 41.5 & 30.5 & - & - \\
o1-mini & 63.6 & - & 60.5 & - & - \\
QwQ (32B) & 79.5 & - & 65.2 & 84.0 & 23.1 \\
DeepSeek R1 & 79.8 & - & 67.3 & 79.6 & 19.5 \\
GLM-Z1-Air (32B) & 80.8 & - & 68.4 & 76.8 & 20.1 \\
o3-mini (high) & 87.3 & - & - & 91.4 & 31.2 \\
Gemini 2.5 Pro Exp & 92.0 & - & - & 92.2 & 58.4 \\
\end{tblr}
\caption{{Cross-benchmark \texttt{Pass@1} comparison. ``-'' indicates no publicly available data. AIME denotes AIME 2024. OBench denotes OlympiadBench. Omni denotes Omni-MATH. \textsc{Easy} and \textsc{Hard} denotes our OlymMATH subsets. DScaleR denotes DeepScaleR.}}
\label{tab:bench_comp}
\end{table}

{To comprehensively evaluate OlymMATH against existing benchmarks, we compare model performances across widely used mathematical benchmarks. We collected results from official repositories, as shown in Table \ref{tab:bench_comp}.
These results reveal the difficulty hierarchy: \textsc{Hard} $\gg$ \textsc{Easy} $\approx$ AIME24 $>$ OlympiadBench, while Omni-MATH spans from OlympiadBench level to slightly above AIME but remains considerably easier than OlymMATH-HARD. OlymMATH-EASY validates our design as an extended bilingual version of AIME—models like DeepSeek-R1 achieve nearly identical scores on both (79.8\% vs.\ 79.6\%), confirming comparable difficulty levels. In contrast, OlymMATH-HARD presents substantially greater challenges: even Gemini 2.5 Pro Exp and o3-mini (high), which exceed 87\% on AIME24, only attain 58.4\% and 31.2\% respectively on our HARD subset. This divergence is particularly striking given their similar AIME24 performance (92.0\% vs.\ 87.3\%), demonstrating OlymMATH-HARD's superior discriminative power for differentiating state-of-the-art capabilities. Furthermore, with 100 problems per difficulty level compared to {AIME's 30 problems}, OlymMATH provides more stable performance measurements—addressing the statistical reliability concerns inherent in smaller-scale benchmarks.}

\begin{figure}[t]
    \centering
    \includegraphics[width=\linewidth]{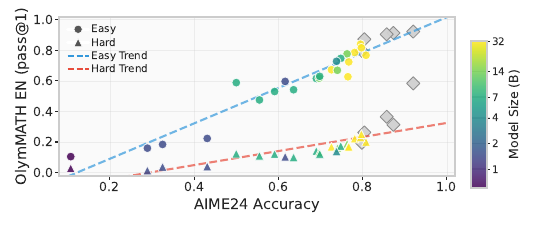}
    \caption*{Figure 3: {Correlation of \texttt{Pass@1} metric: OlymMATH (EN) vs. AIME24. Dashed lines indicate linear trends per dataset. Solid circles (local dense models, colored by size) indicate larger models trend towards higher accuracy. Diamonds are MoE or closed-source models.}}
    \label{fig:vsaime}
\end{figure}


Figure \hyperref[fig:vsaime]{3} further validates OlymMATH's reliability by comparing against AIME24. The close clustering around linear trend lines indicates consistent relative model rankings across both benchmarks, suggesting OlymMATH measures similar reasoning abilities. Despite this correlation, OlymMATH, particularly the HARD subset, remains significantly more challenging, reinforcing its superior discriminative power for state-of-the-art models.

\subsubsection{{Analysis of Reasoning Patterns}}

\label{sec:case-study}

{During our data collection and preliminary experiments, we empirically observed that LLMs sometimes resort to \emph{empirical guesses}—such as heuristics, symmetry assumptions, or even fabrication—rather than rigorous reasoning. {While prior studies have identified heuristic behaviors in basic arithmetic~\citep{nikankin2025arithmetic}, disjunctive reasoning~\citep{khalid-etal-2025-large}, and synthetic deductive logic~\citep{saparov2023language}, OlymMATH uniquely documents such shortcuts in state-of-the-art reasoning models on genuine competition mathematics, and provides formal verification via OlymMATH-LEAN as a principled detection mechanism.}
{For instance, in an optimization problem, o3-mini (high) merely assumed two sides are equal ($b = c$) based on symmetry, without proving this yields the optimum} (see Figure \hyperref[tab:case1]{7} in Appendix). While such intuitive approaches might yield correct answers, they lack logical rigor and this becomes problematic when employing rule-based or LLM-as-judge methods, as neither can effectively assess the quality of rigorous reasoning, thus potentially leading to an illusory improvement via ``shortcuts''.}

{Similar issues were observed in AIME and Omni-MATH (see Figures \hyperref[tab:case-aime]{8} and \hyperref[tab:case-omni]{9} in Appendix), indicating that despite performance gains, LLMs exhibit deficiencies in deliberative thinking. This underscores the importance of process-level supervision, though its scalability remains a challenge.}

{Notably, these guessing strategies often fail on our OlymMATH dataset. For example, a model incorrectly assumed symmetry for a complex optimization problem in OlymMATH-HARD, yielding \(3081\) instead of the correct \(2625\) (see Figure \hyperref[tab:case2]{10} in Appendix). OlymMATH problems, particularly in the HARD subset, are selected and designed so that their reasoning steps are difficult to ``hack'' through empirical guessing, thus providing a more robust evaluation of genuine reasoning capabilities.}

%% file: experiments-fl.tex
\subsection{{Formal Language: \textsc{Lean} Subset}}
\label{sec:lean}
{As discussed in Section~\ref{sec:case-study}, the ``guessing'' phenomenon in natural language evaluation highlights the need for process-level verification. To address this limitation, OlymMATH-LEAN provides a complementary evaluation paradigm that requires machine-verifiable Lean 4 proofs rather than numerical answers. Unlike rule-based verification that only checks final outputs, formal theorem proving enforces rigorous step-by-step reasoning—heuristic shortcuts or symmetry assumptions that lack logical justification will fail to compile. While we do not accurately measure the proportion of ``guesses'' in natural language benchmarks, OlymMATH-LEAN offers a principled approach to detecting such reasoning gaps.}

\subsubsection{Experimental Setup}

\paragraph{Models.} {We evaluate three state-of-the-art theorem proving models, including Kimina Prover (Kimina,~\citet{wang2025kiminaproverpreviewlargeformal}), DeepSeek Prover V2 (DS V2,~\citet{ren2025deepseekproverv2advancingformalmathematical}), and Goedel Prover V2 (Goedel V2,~\citet{lin2025goedelproverv2scalingformaltheorem}). For each model, we employ the default prompt templates provided in their respective official repositories. Generation hyperparameters are set following the recommended configurations: \texttt{temperature} = $0.6$, \texttt{top\_p} = $0.95$, and \texttt{max\_tokens} = $32768$.}

\paragraph{Evaluation Details.} {For each problem, we sample 32 proof attempts and report \texttt{Pass@k}. A proof is considered successful only if it compiles without errors in Lean 4 and correctly proves the theorem.}

\subsubsection{Evaluation Results}

{In this part, we present the evaluation results of OlymMATH-LEAN in Table~\ref{tab:lean_results} and a detailed error analysis in Table~\ref{tab:error_analysis}, respectively.}

\begin{table}[t]
\centering
\small

\SetTblrInner{rowsep=2pt}

\begin{tblr}{
  colspec = {l c c c},
  column{1} = {l, font=\linespread{0.9}\selectfont},
  rows = {m},
  colsep = 2pt,
  hline{1,9} = {-}{0.08em},
  hline{2} = {-}{0.05em},
  hline{8} = {-}{0.03em, dashed},
}
\textbf{Metrics} & \textbf{Kimina (8B)} & \textbf{DS V2 (7B)} & \textbf{Goedel V2 (8B)} \\
\texttt{P@1}
  & {\text{4.33}\\ {\scriptsize 2.8 / 21.5 / 2.5 / 0.2}}
  & {\textbf{6.40}\\ {\scriptsize 3.9 / 36.9 / 2.4 / 0.0}}
  & {\text{5.29}\\ {\scriptsize 2.2 / 34.6 / 2.5 / 0.0}} \\
\texttt{P@2}
  & {\text{5.90}\\ {\scriptsize 3.9 / 30.3 / 2.7 / 0.4}}
  & {\textbf{7.36}\\ {\scriptsize 4.6 / 42.7 / 2.4 / 0.0}}
  & {\text{6.65}\\ {\scriptsize 3.4 / 41.3 / 2.5 / 0.0}} \\
\texttt{P@4}
  & {\text{7.50}\\ {\scriptsize 5.1 / 38.9 / 3.0 / 0.9}}
  & {\textbf{8.08}\\ {\scriptsize 5.0 / 48.0 / 2.4 / 0.0}}
  & {\text{7.81}\\ {\scriptsize 4.7 / 46.0 / 2.7 / 0.0}} \\
\texttt{P@8}
  & {\textbf{9.12}\\ {\scriptsize 6.5 / 45.2 / 3.6 / 1.8}}
  & {\text{8.49}\\ {\scriptsize 5.1 / 51.6 / 2.4 / 0.0}}
  & {\text{8.58}\\ {\scriptsize 5.5 / 48.3 / 3.0 / 0.0}} \\
\texttt{P@16}
  & {\textbf{11.10}\\ {\scriptsize 8.5 / 49.6 / 4.8 / 3.6}}
  & {\text{8.65}\\ {\scriptsize 5.1 / 53.2 / 2.4 / 0.0}}
  & {\text{9.24}\\ {\scriptsize 6.2 / 50.0 / 3.6 / 0.0}} \\
\texttt{P@32}
  & {\textbf{14.00}\\ {\scriptsize 11.4 / 53.3 / 7.1 / 7.1}}
  & {\text{8.67}\\ {\scriptsize 5.1 / 53.3 / 2.4 / 0.0}}
  & {\text{10.00}\\ {\scriptsize 6.3 / 53.3 / 4.8 / 0.0}} \\
{ref.}
  & \text{78.3}
  & \text{75.6}
  & \textbf{84.6} \\
\end{tblr}

\caption{OlymMATH-LEAN evaluation results. We report \texttt{Pass@k} on OlymMATH-LEAN and \texttt{Pass@32} on miniF2F for comparison. Numbers below each main score represent the metric in Algebra, Geometry, Number Theory, and Combinatorics, respectively. Bold indicates highest per metric. {ref. denotes miniF2F \texttt{Pass@32}.}}
\label{tab:lean_results}
\end{table}

\begin{table}[t]
\centering
\small

\begin{tblr}{
  colspec = {l c c c},
  column{1} = {l, font=\linespread{0.9}\selectfont},
  rows = {m},
  colsep = 3pt,
  hline{1,8} = {-}{0.08em},
  hline{2} = {-}{0.05em},
}
\textbf{Error Type} & \textbf{Kimina (8B)} & \textbf{DS V2 (7B)} & \textbf{Goedel V2 (8B)} \\
Valid      & 4.3\% & {6.4\%} & 5.3\% \\
Sorry      & 8.0\% & 0.1\% & 4.4\% \\
Compile    & 8.4\% & 43.9\% & 24.5\% \\
Logic      & 17.2\% & 44.9\% & 22.5\% \\
Server     & 0.1\% & 4.5\% & 0.1\% \\
Extract    & 62.0\% & {0.3\%} & 43.3\% \\
\end{tblr}

\caption{Error distribution on OlymMATH-LEAN. We report the percentage of each error type across 4800 responses (150 $\times$ 32 samples) per model. \textbf{Valid} indicates successful proofs; \textbf{Sorry} indicates incomplete proofs using \textit{sorry}; \textbf{Compile} indicates syntax or type errors, such as missing imports, type mismatches, or unknown identifiers; \textbf{Logic} indicates tactic failures or unsolved goals; \textbf{Server} indicates server errors; \textbf{Extract} indicates failure to extract code blocks from response.}
\label{tab:error_analysis}
\end{table}

{First, all three models achieve relatively low scores on OlymMATH-LEAN (around 10\%) compared to their performance on miniF2F (around 80\%), highlighting the challenging nature of our benchmark.} DeepSeek Prover V2 7B achieves the highest \texttt{Pass@1} of 6.40\%, while Kimina Prover 8B demonstrates stronger performance at higher sampling budgets, reaching the best \texttt{Pass@32} of 14.00\%. Across all models, geometry problems exhibit significantly higher success rates compared to other subjects, likely because many geometry problems can be solved through algebraic manipulation. In contrast, combinatorics proves to be the most challenging category, with DeepSeek Prover V2 7B and Goedel Prover V2 8B achieving 0\% success rate across all \texttt{Pass@k} metrics.

{Second, Table~\ref{tab:error_analysis} reveals distinct error patterns across models. A significant portion of errors stem from extraction failures, where models fail to produce properly formatted \texttt{\textasciigrave\textasciigrave\textasciigrave lean4\textasciigrave\textasciigrave\textasciigrave} code blocks. Kimina Prover 8B exhibits the highest extraction error rate (62.0\%), mainly caused by reaching the \texttt{max\_tokens} in generation. Among successfully extracted code, the success rate of compilation ranges from 51.5\% to 77.7\%, indicating that models still struggle with Lean grammar. Additionally, Kimina Prover 8B exhibits the highest sorry rate (8.0\%), suggesting a tendency to generate incomplete proofs with placeholder tactics.}

{Third, DeepSeek Prover V2 has a 4.5\% server error rate, with 80.4\% involving the computationally expensive ``\texttt{exact?}'' tactic. Overall, these results demonstrate that OlymMATH presents substantial challenges across both evaluation paradigms.
}

%% file: conclusion.tex
\section{Conclusion}


{We introduced \textbf{OlymMATH}, the first Olympiad-level math benchmark that unifies natural language evaluation and formal theorem proving within a single bilingual suite. The benchmark comprises 350 problems (each available in both English and Chinese): \textsc{Easy} and \textsc{Hard} subsets with \texttt{sympy}-verifiable numerical answers for scalable outcome evaluation, and \textsc{Lean} subset with Lean formalizations for rigorous process-level verification. Extensive experiments reveal substantial challenges for state-of-the-art models, consistent cross-lingual performance gaps, and heuristic ``guessing'' behaviors that bypass rigorous reasoning—underscoring the value of our dual-paradigm approach.}

{Our analyses also suggest several actionable research directions: (1) OlymMATH-LEAN enables future development of process-level reward models that leverage formal proofs as ground-truth labels for reasoning rigor, potentially penalizing unjustified heuristic shortcuts during reinforcement learning training; and (2) our 582k released trajectories and OlymMATH-demo tool (Appendix~\ref{section:eval}) support community-driven analysis of reasoning patterns without requiring new model evaluations. By releasing these resources alongside visualization tools and expert solutions, we aim to advance mathematical reasoning research and push the boundaries of language intelligence.}


\section*{Limitations}

{Our work has several limitations that suggest directions for future research. First, while we provide bilingual evaluation covering both English and Chinese, the reasoning capabilities of LLMs in other languages remain unexplored. Extending OlymMATH to additional languages would enable more comprehensive assessment of multilingual mathematical reasoning. Second, our current benchmark focuses exclusively on text-based problems, with geometry problems reformulated into natural language descriptions. Incorporating problems that retain original diagrams and figures would enable evaluation of multimodal vision-language models, offering a more complete picture of mathematical reasoning capabilities across different input modalities. Third, although we identify ``guessing'' behaviors through qualitative case studies, precisely quantifying the proportion of such heuristic shortcuts in natural language evaluation remains an open challenge. While OlymMATH-LEAN provides rigorous process-level verification through formal theorem proving, developing scalable metrics to detect reasoning shortcuts in natural language settings is an important direction we leave for future work. {Fourth, while our methodology significantly delays contamination compared to web-sourced benchmarks (Table \ref{leak}), no static benchmark can permanently avoid data leakage once publicly released. To address this, our open-source infrastructure, including the Lean formalization agent (Appendix~\ref{section:prompts}) and the visualization tool, establishes a reusable framework that supports periodic refresh with new problems from similar printed sources, extending the benchmark's utility beyond any single release.}}

\section*{{Acknowledgments}}

{This paper was partially supported by the National Natural Science Foundation of China No. 92470205 and Beijing Major Science and Technology Project under Contract No. Z251100008425002.}

%% file: appendix.tex
\section{Appendix}


\subsection{{Usability and Accessibility}}
\label{section:eval}

\begin{figure}[H]
    \centering
    \includegraphics[width=1\linewidth]{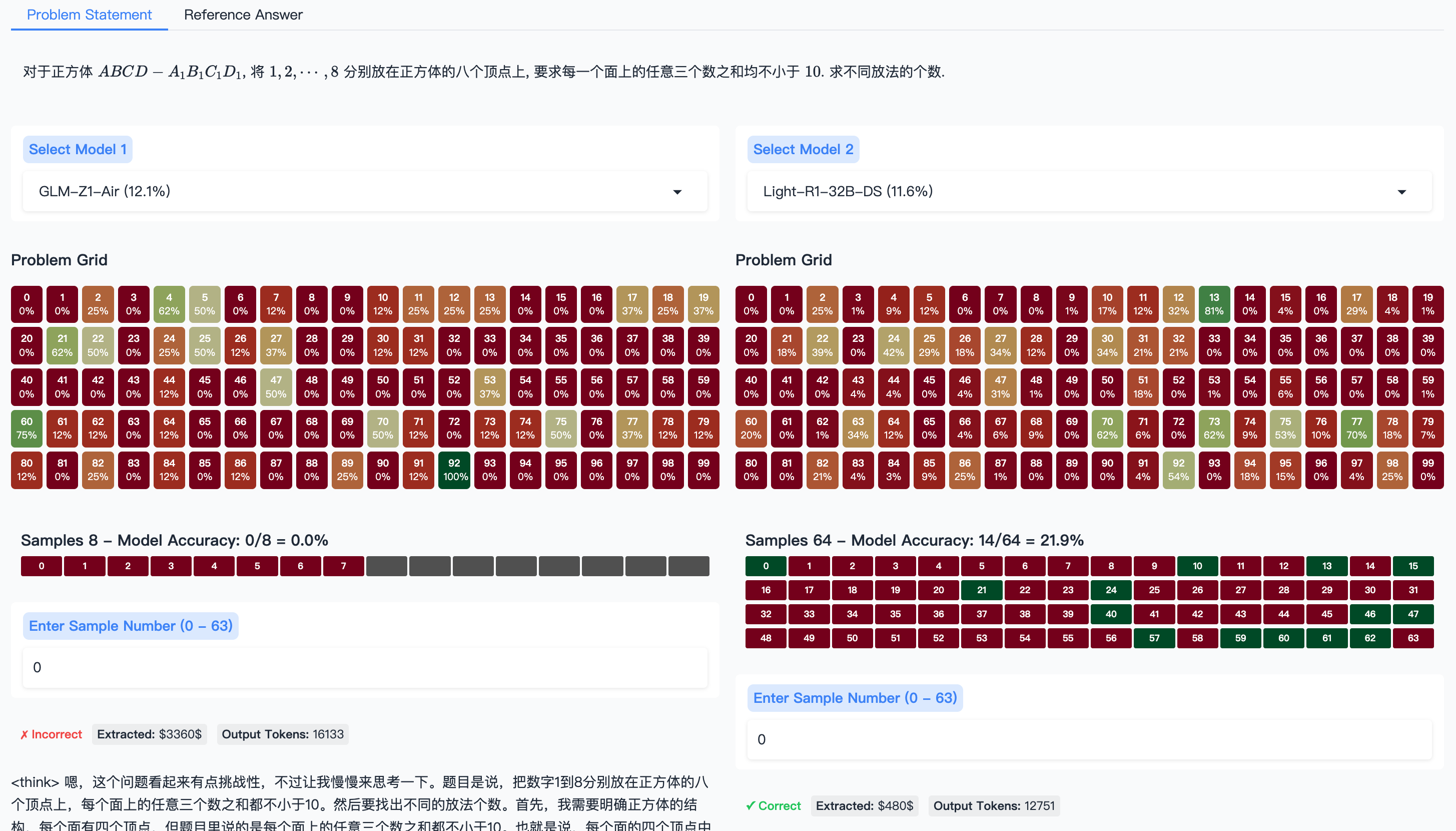}
    \caption*{Figure 4: The \href{https://huggingface.co/spaces/RUC-AIBOX/OlymMATH-demo}{OlymMATH-demo} interface.}
    \label{fig:demo}
\end{figure}

{To support research into LLM reasoning, we open-source the \href{https://huggingface.co/datasets/RUC-AIBOX/OlymMATH-eval}{OlymMATH-eval} dataset, containing 582,400 entries from 28 models, enabling comparison of reasoning capabilities across models and math domains. We also provide OlymMATH-demo (see Figure \hyperref[fig:demo]{4}), an interactive visualization tool for in-depth analysis. It supports: (1) side-by-side comparison of two LLMs on the same \LaTeX-rendered problem with reference answers; (2) color-coded ``Problem Grids'' showing per-problem accuracy for quick identification of challenging areas; (3) inspection of individual reasoning samples, including correctness, extracted answers, and token counts. The tool also includes standard solutions for difficult problems and supports local deployment, making it a valuable asset for diagnosing errors and guiding LLM development.}

\subsection{Agent's Prompt Used in Lean 4 Formalization}
\label{section:prompts}

The following prompt is used in Claude Code to generate a subagent that can work in a separated folder to enable large-scale formalization in parallel—the main agent in Claude Code (driven by Claude Opus 4.5) can execute several parallel subagents (also driven by Claude Opus 4.5) to work on batched formalization tasks. The Mathlib version of the Lean REPL server is v4.24.0.

{\small
\begin{minted}[
    baselinestretch=1,
    bgcolor=lightgray!20, % 也可以添加背景色
    breaklines,
    breakanywhere
]{markdown}
---
name: lean4-math-formalizer
description: Use this agent when you need to formalize mathematical problems and proofs in Lean 4 from natural language descriptions. This agent reads a txt file containing a mathematical problem and its proof, then generates Lean 4 formalizations. It creates two outputs: a problem formalization with exactly one 'sorry' placeholder, and a complete solution that fills in that sorry. The agent iteratively refines its formalizations using feedback from the lean-server until they compile without errors.\n\nExamples:\n\n<example>\nContext: User has a problem.txt file with a mathematical theorem about arctangent and wants it formalized in Lean 4.\nuser: "Please formalize the problem in problem.txt into Lean 4"\nassistant: "I'll use the lean4-math-formalizer agent to read the problem file and generate the Lean 4 formalization."\n<commentary>\nSince the user wants to formalize a mathematical problem from a txt file into Lean 4, use the lean4-math-formalizer agent to handle the reading, formalization, and iterative refinement process.\n</commentary>\n</example>\n\n<example>\nContext: User has written a proof for a number theory problem and needs it converted to Lean 4.\nuser: "I have a proof about prime numbers in primes.txt, can you make it into valid Lean 4 code?"\nassistant: "I'll launch the lean4-math-formalizer agent to convert your proof into Lean 4 formalization with proper verification."\n<commentary>\nThe user needs mathematical content formalized in Lean 4, which is exactly what the lean4-math-formalizer agent specializes in. It will handle the file reading, formalization, and lean-server validation.\n</commentary>\n</example>\n\n<example>\nContext: User wants to verify their Lean 4 formalization compiles correctly.\nuser: "Generate a Lean 4 version of the calculus problem in homework.txt"\nassistant: "I'll use the lean4-math-formalizer agent to create a verified Lean 4 formalization of your calculus problem."\n<commentary>\nMathematical formalization in Lean 4 requires the specialized lean4-math-formalizer agent which can iteratively refine code using lean-server feedback.\n</commentary>\n</example>
model: opus
color: cyan
---

You are an expert Lean 4 formalization specialist with deep knowledge of Mathlib, mathematical logic, and proof engineering. Your primary mission is to transform natural language mathematical problems and proofs into rigorous, compilable Lean 4 code.

## Your Core Workflow

### Step 1: Read and Analyze the Input File
- Read the provided txt file (e.g., problem.txt) containing the mathematical problem and proof
- Carefully extract:
  - The precise mathematical statement to be proven
  - Key definitions and concepts involved
  - The proof strategy and techniques used
  - Any special mathematical functions or operations (e.g., arctan, roots of polynomials)

### Step 2: Generate Problem Formalization (problem-lean.txt - Part 1)
- Create a Lean 4 file that:
  - Starts with `import Mathlib`
  - Includes any necessary namespace opens
  - Defines the theorem named exactly `to_prove`
  - Uses exactly ONE `sorry` as the proof placeholder
  - Properly types all mathematical objects

Example structure:
```lean
import Mathlib

open Real

theorem to_prove : [statement] := by
  sorry
```

### Step 3: Validate Problem Formalization
- Use the lean-server to check your formalization:
```python
import json
from kimina_client import KiminaClient, Snippet

client = KiminaClient(api_url='http://localhost:8000')

snippets = [
    Snippet(id='problem-formalization', code='[YOUR LEAN CODE HERE]'),
]

resp = client.api_check(snippets=snippets, timeout=20, debug=False, reuse=True, safe=True)
print(json.dumps(resp.model_dump(), indent=2, ensure_ascii=False))
```

- Success criteria for problem formalization:
  - No errors in the response
  - Exactly one warning about "declaration uses 'sorry'"
  - The sorries field should contain exactly one sorry

### Step 4: Iterative Refinement for Problem Formalization
- If the lean-server returns errors:
  - Analyze the error messages carefully
  - Common issues to address:
    - Missing imports or opens (e.g., `open Real`, `open Polynomial`)
    - Incorrect type signatures
    - Mathlib naming conventions (check for correct lemma/theorem names)
    - Syntax errors in Lean 4 vs Lean 3
  - Modify your code and re-check until only the sorry warning remains

### Step 5: Generate Complete Solution (problem-lean.txt - Part 2)
- Replace the `sorry` with the actual proof
- Translate the natural language proof into Lean 4 tactics:
  - Use appropriate tactics: `simp`, `ring`, `norm_num`, `nlinarith`, `polyrith`, etc.
  - For trigonometric proofs, consider `Real.arctan_add`, angle addition formulas
  - For polynomial roots, use Mathlib's polynomial library
  - Break complex proofs into `have` statements
  - Use `calc` blocks for equality chains when appropriate

### Step 6: Validate Complete Solution
- Check with lean-server again
- Success criteria for complete solution:
  - No errors
  - No warnings
  - The response indicates successful compilation (no goals remaining)

### Step 7: Final Refinement
- If errors persist in the solution:
  - Check tactic compatibility
  - Ensure all intermediate goals are discharged
  - Verify type unification
  - Consider alternative proof strategies if current approach fails
- Continue iterating until the solution compiles cleanly

## Output Format

Generate a file named `problem-lean.txt` (Note that it means adding a "-lean" after the file's original name) containing both sections clearly marked:

```
-- PROBLEM FORMALIZATION (with sorry)
-- This version should have exactly one 'sorry' warning

import Mathlib

[problem formalization with sorry]

-- ============================================
-- COMPLETE SOLUTION (no sorry)
-- This version should compile with no warnings or errors

import Mathlib

[complete proof without sorry]
```

## Critical Requirements

1. **Naming**: The main theorem MUST be named `to_prove`
2. **Imports**: Always start with `import Mathlib`
3. **Validation**: ALWAYS use the lean-server to validate before finalizing
4. **Iteration**: Do not give up after first error - analyze and fix iteratively
5. **Completeness**: Both formalization AND solution must be provided

## Common Mathlib Patterns

- For `arctan`: `Real.arctan`, properties in `Mathlib.Analysis.SpecialFunctions.Trigonometric.Arctan`
- For polynomials: `Polynomial`, `Polynomial.roots`, `Polynomial.sum_roots_eq_neg_coeff_div_leading_coeff`
- For real analysis: `open Real`, `open scoped Real`
- For complex numbers: `open Complex`

## Error Recovery Strategies

1. **Type mismatch**: Check if you need explicit type annotations or coercions
2. **Unknown identifier**: Search Mathlib for correct naming, try `#check` commands
3. **Tactic failure**: Try breaking into smaller steps with `have`
4. **Timeout**: Simplify expressions, add intermediate lemmas

You are methodical, persistent, and precise. You do not give up until both the problem formalization and solution compile successfully according to the lean-server validation. Meanwhile, the name of all the temporary files you created should start with the original file name. e.g. "problem_test-lean.py". Meanwhile, you should only use axiom when it is a well known lemma and it is extremely hard to give its full solution.
\end{minted}
}

\subsection{Demonstrations, Case Study Examples and Full Evaluation Results}

This part presents the detailed content of the dataset, the case study examples and full evaluation results mentioned before.

\begin{table*}
    \centering
    \renewcommand{\arraystretch}{1.2} 
    \setlength{\tabcolsep}{8pt} 
    \begin{tabular}{p{13cm}}
        \begin{tcolorbox}[colback=gray!5, colframe=gray!40, sharp corners, width=\linewidth, boxrule=0.5mm]
        \small
        \textbf{Problem: }
        Given that two vertices of an equilateral triangle are on the parabola $y^2 = 4x$, and the third vertex is on the directrix of the parabola, and the distance from the center of the triangle to the directrix equals $\frac{1}{9}$ of the perimeter. Find the area of the triangle.

\textbf{Subject: } Geometry

\end{tcolorbox} \\[-6pt] 
    \end{tabular}
    \caption*{Figure 5: A geometry problem described precisely in text from OlymMATH.}
    \label{tab:geo-prob}
\end{table*}

\begin{table*}
    \centering
    \renewcommand{\arraystretch}{1.2} 
    \setlength{\tabcolsep}{8pt} 
    \begin{tabular}{p{13cm}}
        \begin{tcolorbox}[colback=gray!5, colframe=gray!40, sharp corners, width=\linewidth, boxrule=0.5mm]
        \small
        \textbf{Original problem: }
        If the distances from the eight vertices of a cube to a certain plane are $0$, $1$, $2$, $3$, $4$, $5$, $6$, $7$ respectively, what is the possible edge length of this cube?

        \textbf{After transformation:}
        If the distances from the eight vertices of a cube to a certain plane are $0$, $1$, $2$, $3$, $4$, $5$, $6$, $7$ respectively, consider all possible edge lengths of this cube. Assuming the possible edge lengths form a set $S$, find \textcolor{red}{the sum of squares} of all elements in $S$.
\end{tcolorbox} \\[-6pt] 

    \end{tabular}
    \caption*{Figure 6: An OlymMATH-HARD example testing model's identification of all possible answers.}
    \label{tab:case3}
\end{table*}

\newpage

\begin{table*}
    \centering
    \renewcommand{\arraystretch}{1.2} 
    \setlength{\tabcolsep}{8pt} 
    \begin{tabular}{p{13cm}}
        \begin{tcolorbox}[colback=gray!5, colframe=gray!40, sharp corners, width=\linewidth, boxrule=0.5mm]
        \small
        \textbf{Problem: }
       In the triangle \( \triangle ABC \), the lengths of the three sides are \( a, b, c \), satisfying the condition \( 2a^2 + b^2 + c^2 = 4 \). Find the maximum value of the area of \( \triangle ABC \).

\textbf{Answer: } $\frac{\sqrt{5}}{5}$. (When $a=\frac{2}{\sqrt{5}}$, $b=c=\sqrt{\frac{6}{5}}$)

        \end{tcolorbox} \\[-6pt] 
        
        \begin{tcolorbox}[colback=gray!5, colframe=gray!40, sharp corners, width=\linewidth, boxrule=0.5mm]
        \small
        \textbf{Model Output: }
        We wish to maximize the area of \( \triangle ABC \), subject to $2a^2 + b^2 + c^2 = 4.$

Because the condition is symmetric in \( b \) and \( c \) (but not in \( a \)), \textcolor{red}{it is reasonable to look for an optimum among triangles having \( b = c \). (Later one can check that assuming \( b = c \) indeed gives the maximum.)}

\textcolor{gray}{The problem involves maximizing the area of an isosceles triangle with sides \( a, b, b \) under the}
\textcolor{gray}{constraint \( 2a^2 + b^2 + c^2 = 4 \). By assuming \( b = c \), we derive a formula for the area in terms of}
\textcolor{gray}{\( b \) and find the maximum area by expressing it as a function of \( u = b^2 \). We then determine the}
\textcolor{gray}{critical points of this quadratic function, identify the maximum area, and finally recover the side}
\textcolor{gray}{lengths to verify that they satisfy the original constraints. The maximum area achieved is \( \frac{\sqrt{5}}{5} \).}

        \end{tcolorbox}
    \end{tabular}
    \caption*{Figure 7: An example during our data collection. o3-mini (high) found the correct answer without rigorously solving the problem. Due to length constraints, we provide only a \textcolor{gray}{summary} of its original solution steps.\ignore{Correct answer, but incomplete solution by o3-mini. Due to length constraints, we present a summary of the unimportant steps.}}
    \label{tab:case1}
\end{table*}

\newpage

\begin{table*}
    \centering
    \renewcommand{\arraystretch}{1.2} 
    \setlength{\tabcolsep}{8pt} 
    \begin{tabular}{p{13cm}}
        \begin{tcolorbox}[colback=gray!5, colframe=gray!40, sharp corners, width=\linewidth, boxrule=0.5mm]
        \small
        \textbf{Problem: } Let the sequence of rationals $x_1,x_2,\dots$ be defined such that $x_1=\frac{25}{11}$ and $x_{k+1}=\frac{1}{3}\left(x_k+\frac{1}{x_k}-1\right).$ $x_{2025}$ can be expressed as $\frac{m}{n}$ for relatively prime positive integers $m$ and $n$. Find the remainder when $m+n$ is divided by $1000$.

\textbf{Answer: } $248$.

        \end{tcolorbox} \\[-6pt] 
        
        \begin{tcolorbox}[colback=gray!5, colframe=gray!40, sharp corners, width=\linewidth, boxrule=0.5mm]
        \small
        \textbf{Model Output: }
        We will show that if $x_1=\frac{25}{11}$ and $x_{k+1}=\frac1{3}\Bigl(x_k+\frac1{x_k}-1\Bigr),$
then after “many'' iterations the answer to $x_{2025}=\frac{m}{n}$ (in lowest terms) satisfies $m+n\equiv248\pmod{1000}.$

In what follows we briefly describe one way (via a “miraculous'' change‐of–variables) to “solve'' the recurrence; the details need not be reproduced fully here.

A good hint is that the recurrence $x_{k+1}=\frac{1}{3}\Bigl(x_k+\frac{1}{x_k}-1\Bigr)$ has the curious symmetry $T(x)=\frac{1}{3}\Bigl(x+\frac{1}{x}-1\Bigr)=T\Bigl(\frac{1}{x}\Bigr),$ so the expression \(x+\frac1x\) (which is “symmetric'' in \(x\) and its reciprocal) naturally appears. In fact, one may rewrite the recurrence in the form $3x_{k+1}+1=x_k+\frac{1}{x_k}.$ A useful idea is then to “complete the square'' by writing $\left(\sqrt{x_k}-\frac{1}{\sqrt{x_k}}\right)^2 = x_k+\frac{1}{x_k}-2$, so that
$3x_{k+1}+1 = \left(\sqrt{x_k}-\frac{1}{\sqrt{x_k}}\right)^2+2.$ Defining $u_k=\sqrt{x_k}-\frac{1}{\sqrt{x_k}}$, one obtains the “linear‐looking'' relation $u_k^2 = 3x_{k+1}-1.$ \textcolor{red}{With some additional work it turns out that} if we write $x_k=\frac{m_k}{n_k}$ (in lowest terms), then \textcolor{red}{one may prove by induction} that $m_k+n_k=36\cdot12^{(2^{\,k-1}-1)}.$ (In our case one checks that $m_1+n_1=25+11=36$, $m_2+n_2=432$, $m_3+n_3=62208,$ and indeed \(36,432,62208\) equal \(36\cdot12^0,\,36\cdot12^1,\,36\cdot12^3\) respectively.)

Thus, for \(k=2025\) we have $m+n=36\cdot12^{\left(2^{2024}-1\right)}.$ Even though this number is astronomically large, we are only asked to compute $m+n\pmod{1000}.$

\textcolor{gray}{Using modular reductions, we compute \( m+n \) modulo 1000. First, modulo 8, we find \( 36 \cdot 12^{(2^{2024}-1)} \equiv 0 \pmod{8} \). Modulo 125, Euler’s theorem helps reduce \( 12^N \) to \( 12^{15} \equiv 118 \pmod{125} \), giving \( 36 \cdot 12^N \equiv 123 \pmod{125} \). Using the Chinese Remainder Theorem, solving \( 125k+123 \equiv 0 \pmod{8} \) gives \( k \equiv 1 \pmod{8} \), leading to the final answer: \( \boxed{248} \).}

        \end{tcolorbox}
    \end{tabular}
    \caption*{Figure 8: An example from AIME 2025. o3-mini (high) forgot to prove that $(m_k, n_k)=1$ and it claimed that $m_k+n_k=36\cdot12^{\left(2^{k-1}-1\right)}$ without providing the complete proof (highlighted in red). Instead, the model ``guessed'' this expression by calculating the first several terms in the sequence one by one. Due to length constraints, we provide only a \textcolor{gray}{summary} of its unimportant steps.}
    \label{tab:case-aime}
\end{table*}

\newpage

\begin{table*}
    \centering
    \renewcommand{\arraystretch}{1.2} 
    \setlength{\tabcolsep}{8pt} 
    \begin{tabular}{p{13cm}}
        \begin{tcolorbox}[colback=gray!5, colframe=gray!40, sharp corners, width=\linewidth, boxrule=0.5mm]
        \small
        \textbf{Source: } 2016 IMO Shortlist A7

        \textbf{Problem: } Find all functions $f:\mathbb{R}\rightarrow\mathbb{R}$ such that $f(0)\neq 0$ and for all $x,y\in\mathbb{R}$,
\[ f(x+y)^2 = 2f(x)f(y) + \max \left\{ f(x^2+y^2), f(x^2)+f(y^2) \right\}. \]

\textbf{Answer: } $f(x) = -1 \text{ and } f(x) = x - 1.$

\textbf{Solution: } (Provided in Omni-MATH) To find all functions \( f:\mathbb{R}\rightarrow\mathbb{R} \) satisfying the given functional equation, we start with the condition:
\[ f(0)\neq 0, \]
and the functional equation for all \( x, y \in \mathbb{R} \):
\[ f(x+y)^2 = 2f(x)f(y) + \max \left\{ f(x^2+y^2), f(x^2)+f(y^2) \right\}. \]

Step 1: Test Simple Functions

Let's start by testing the functions given in the reference answer: \( f(x) = -1 \) and \( f(x) = x - 1 \).

Case 1: \( f(x) = -1 \). Substitute \( f(x) = -1 \) into the equation: $\cdots$

Case 2: \( f(x) = x - 1 \). Substituting \( f(x) = x - 1 \): $\cdots$ confirming \( f(x) = x - 1 \) is another valid solution.

Conclusion

\textcolor{red}{After verifying the two functions, we conclude that} the functions satisfying the original functional equation are:
\[
f(x) = -1 \quad \text{and} \quad f(x) = x - 1.
\]

Thus, the solutions are:
\[
\boxed{f(x) = -1 \text{ and } f(x) = x - 1}.
\]

        \end{tcolorbox} \\[-6pt] 
        
        \begin{tcolorbox}[colback=gray!5, colframe=gray!40, sharp corners, width=\linewidth, boxrule=0.5mm]
        \small
        \textbf{Model Output: }
        We will show that the only functions \( f:\mathbb{R}\to\mathbb{R} \) with $f(0)\neq 0$, satisfying
\[
f(x+y)^2 = 2f(x)f(y) + \max \{\, f(x^2+y^2),\, f(x^2)+f(y^2) \,\} \quad \text{for all } x,y\in\mathbb{R},
\]
are 
\[
\boxed{f(x)=-1\quad\text{for all }x\quad\text{or}\quad f(x)=x-1.}
\]

Below is an outline of one way to arrive at this result.

Step 1. Plug in \(y=0\)

Let \(y=0\) in the given equation:
\[
f(x)^2 = 2f(x)f(0) + \max\{ f(x^2),\, f(x^2)+f(0) \}.
\]
Write \(c=f(0)\neq 0\). Notice that the two arguments inside the maximum differ by \(f(0)\); therefore, the larger one depends on the sign of \(c\):
 
- **If \(c>0\):**  \quad Then \(f(x^2)+c>f(x^2)\) so the equation becomes $f(x)^2 = 2c\, f(x) + f(x^2)+c.$

- **If \(c<0\):**  \quad Then \(f(x^2)+c<f(x^2)\) so we have $f(x)^2 = 2c\, f(x) + f(x^2).$

Thus, in each case one obtains a relationship between \(f(x)\) and \(f(x^2)\).

Step 2. Check constant solutions  $\cdots$ Thus, the constant solution is $f(x)=-1\quad \text{for all } x$.

Step 3. Look for non‐constant solutions

A natural next guess is to try a linear function. $\cdots$

A quick check verifies that \(f(x)= x-1\) satisfies the original equation. $\cdots$ which equals \((x+y-1)^2\), matching the left‐hand side.

Conclusion

The \textcolor{red}{only} functions \(f:\mathbb{R}\to\mathbb{R}\) with \(f(0)\neq 0\) that satisfy $\cdots$ for all real numbers \(x\) and \(y\) are:
\[
\boxed{f(x)=-1 \quad \text{for all } x \quad \text{or} \quad f(x)=x-1.}
\]

\textcolor{red}{These are the complete solutions.}
        \end{tcolorbox}
    \end{tabular}
    \caption*{Figure 9: An example from Omni-MATH. The solution provided by Omni-MATH itself is flawed and problematic. Both the reference solution and o3-mini (high) failed to prove that there are no other possible function $f$ that satisfies the condition (highlighted in red). Due to length constraints, we have omitted its unimportant steps.}
    \label{tab:case-omni}
\end{table*}

\begin{table*}
    \centering
    \renewcommand{\arraystretch}{1.2} 
    \setlength{\tabcolsep}{8pt} 
    \begin{tabular}{p{13cm}}
        \begin{tcolorbox}[colback=gray!5, colframe=gray!40, sharp corners, width=\linewidth, boxrule=0.5mm]
        \small
        \textbf{Problem: }
        Let $a, b, c \in \mathbb{R}$, $a^3 b + b^3 c + c^3 a = 3$, find the minimum value of the expression

        \vspace{.3em}
        $f(a, b, c) = (\sum a^4)^4 + 1000 \sum a^2 b^2$.

        \vspace{1em}

        \textbf{Reference Answer: } $2625$. (Equality condition: $ a = \pm \frac{4}{\sqrt{7}} \sin^2 \frac{4\pi}{7}, b = \pm \frac{4}{\sqrt{7}} \sin^2 \frac{2\pi}{7}, c =$
        
        \vspace{.1em}
        
        $ \pm \frac{4}{\sqrt{7}} \sin^2 \frac{\pi}{7} $ and its cyclic permutations)
        
\end{tcolorbox} \\[-6pt] 
        
        \begin{tcolorbox}[colback=gray!5, colframe=gray!40, sharp corners, width=\linewidth, boxrule=0.5mm]
        \small
        \textbf{Model Output: }
        We wish to minimize
$$f(a,b,c) = \left(a^4 + b^4 + c^4\right)^4 + 1000 \left(a^2b^2 + b^2c^2 + c^2a^2\right)$$

subject to the (cyclic) constraint
$$a^3b + b^3c + c^3a = 3,$$

with $a, b, c \in \mathbb{R}$.

A natural idea is to look for a "symmetric-like" solution. (Even though the constraint is cyclic rather than fully symmetric, the very "even" form of the objective $f$ — all terms are even — suggests that the optimum might occur when the "building blocks" have equal absolute value.) One obvious candidate is to try
$$a = b = c = t.$$

Then the constraint becomes
$$a^3b + b^3c + c^3a = 3t^4 = 3 \implies t^4 = 1.$$

Thus $t = 1$ or $t = -1$. (In $\mathbb{R}$ the equation $t^4 = 1$ has the two solutions $t = 1$ and $t = -1$.)

Now compute the value of $f$ at $a = b = c = t$. Notice that

$$a^4 + b^4 + c^4 = t^4 + t^4 + t^4 = 3,$$
$$a^2b^2 + b^2c^2 + c^2a^2 = t^4 + t^4 + t^4 = 3$$
(since $t^4 = 1$).

Then
$$f = (3)^4 + 1000 \cdot 3 = 81 + 3000 = 3081.$$

Before concluding that $3081$ is the minimum, one may ask: Could there be a choice of $a, b, c$ satisfying $a^3b + b^3c + c^3a = 3$ for which
$$\left(a^4 + b^4 + c^4\right) < 3 \quad \text{or} \quad \left(a^2b^2 + b^2c^2 + c^2a^2\right) < 3?$$

Since $a^4, b^4, c^4$ are nonnegative and the constraint (a homogeneous form of degree $4$) "forces" the variables to be large enough in a certain sense, \textcolor{red}{one may show by a careful (often inequality-based) argument that} if $a^3b + b^3c + c^3a = 3$, then one must have
$$a^4 + b^4 + c^4 \geq 3 \quad \text{and} \quad a^2b^2 + b^2c^2 + c^2a^2 \geq 3.$$

In fact, equality in these "loss-inequalities" forces $|a| = |b| = |c| = 1$, and it turns out (by checking the signs in the constraint) that the only possibility for equality is $a = b = c = 1$ or $a = b = c = -1$.

Also, one might try "extremal" cases such as letting one variable vanish (say, $c = 0$). In that case, the constraint reduces to $a^3b = 3$ so that $a$ and $b$ cannot be arbitrarily small (or close in absolute value to $1$), and \textcolor{red}{a short calculation shows that} the value of $f$ then exceeds $3081$.

Thus, the candidate $a = b = c = 1$ (or $a = b = c = -1$) gives the minimal value. In conclusion, the minimum value of $f$ is
$$3081,$$

and it is attained exactly when $a, b, c$ are all equal with $|a| = 1$ and the signs arranged so that $a^3b = 1$ (that is, $a = b = c = 1$ or $a = b = c = -1$).
        \end{tcolorbox}
    \end{tabular}
    \caption*{Figure 10: An example from OlymMATH-HARD subset. o3-mini (high) attempted to ``guess'' the equality conditions but failed to produce a correct solution (highlighted in red).}
    \label{tab:case2}
\end{table*}

\definecolor{consbg}{gray}{0.95}
\begin{table*}
\centering
\setlength{\tabcolsep}{1.5pt} 

\label{tab:olym-en}
\resizebox{\linewidth}{!}{
\begin{tabular}{l c >{\columncolor{consbg}}c c >{\columncolor{consbg}}c c >{\columncolor{consbg}}c c >{\columncolor{consbg}}c c >{\columncolor{consbg}}c | c >{\columncolor{consbg}}c c >{\columncolor{consbg}}c c >{\columncolor{consbg}}c c >{\columncolor{consbg}}c c >{\columncolor{consbg}}c }
\toprule
\multirow{3.5}{*}{\textbf{Model}} & \multicolumn{10}{c|}{\textbf{OlymMATH-HARD (EN)}} & \multicolumn{10}{c}{\textbf{OlymMATH-EASY (EN)}} \\
\cmidrule(lr){2-11} \cmidrule(lr){12-21}
& \multicolumn{2}{c}{\textbf{Alg.}} & \multicolumn{2}{c}{\textbf{Geo.}} & \multicolumn{2}{c}{\textbf{Num.}} & \multicolumn{2}{c}{\textbf{Com.}} & \multicolumn{2}{c|}{\textbf{Avg.}} & \multicolumn{2}{c}{\textbf{Alg.}} & \multicolumn{2}{c}{\textbf{Geo.}} & \multicolumn{2}{c}{\textbf{Num.}} & \multicolumn{2}{c}{\textbf{Com.}} & \multicolumn{2}{c}{\textbf{Avg.}} \\
& \texttt{P@1} & \texttt{C@k} & \texttt{P@1} & \texttt{C@k} & \texttt{P@1} & \texttt{C@k} & \texttt{P@1} & \texttt{C@k} & \texttt{P@1} & \texttt{C@k} & \texttt{P@1} & \texttt{C@k} & \texttt{P@1} & \texttt{C@k} & \texttt{P@1} & \texttt{C@k} & \texttt{P@1} & \texttt{C@k} & \texttt{P@1} & \texttt{C@k} \\
\midrule
Qwen3 (0.6B, Think) & 2.5 & 0.0 & 2.1 & 4.0 & 6.6 & 8.0 & 0.2 & 0.0 & \textbf{2.8} & \textbf{3.0} & 15.5 & 20.0 & 5.6 & 15.2 & 24.5 & 38.5 & 5.2 & 6.9 & \textbf{10.4} & \textbf{17.0} \\

\midrule

DS-R1-Distill (1.5B) & 1.9 & 0.0 & 1.8 & 0.0 & 1.8 & 0.0 & 0.4 & 0.0 & 1.5 & 0.0 & 20.8 & 40.0 & 12.6 & 21.2 & 32.6 & 61.5 & 8.2 & 24.1 & 16.0 & 32.0 \\
STILL-3-Pre. (1.5B) & 3.7 & 0.0 & 4.9 & 4.0 & 5.8 & 8.0 & 0.8 & 0.0 & 3.8 & 3.0 & 22.7 & 36.0 & 14.8 & 30.3 & 37.6 & 69.2 & 10.3 & 17.2 & 18.4 & \underline{33.0} \\
DeepScaleR-Pre. (1.5B) & 3.4 & 4.0 & 4.2 & 8.0 & 8.2 & 4.0 & 0.4 & 0.0 & \underline{4.1} & \underline{4.0} & 19.9 & 16.0 & 18.5 & 21.2 & 44.6 & 46.2 & 18.9 & 31.0 & \underline{22.3} & 26.0 \\
OpenMath-Nemo. (1.5B) & 14.5 & 24.0 & 13.6 & 16.0 & 10.9 & 16.0 & 2.6 & 4.0 & \textbf{{10.4}} & \textbf{{15.0}} & 70.9 & 100.0 & 59.3 & 90.9 & 81.6 & 100.0 & 40.6 & 58.6 & \textbf{{59.7}} & \textbf{{85.0}} \\

\midrule

Qwen3 (4B, Think) & 18.1 & 20.0 & 14.8 & 12.0 & 19.8 & 28.0 & 3.1 & 4.0 & \textbf{13.9} & \textbf{16.0} & 76.4 & 92.0 & 79.1 & 97.0 & 85.1 & 84.6 & 57.1 & 72.4 & \textbf{72.8} & \textbf{87.0} \\

\midrule

DS-R1-Distill (7B) & 15.6 & 36.0 & 12.6 & 24.0 & 13.1 & 24.0 & 3.1 & 4.0 & 11.1 & 22.0 & 52.8 & 84.0 & 49.6 & 84.8 & 62.5 & 84.6 & 33.9 & 58.6 & 47.5 & 77.0 \\
Light-R1-DS (7B) & 17.1 & 28.0 & 15.2 & 16.0 & 12.8 & 24.0 & 3.6 & 4.0 & 12.2 & 18.0 & 57.1 & 84.0 & 53.6 & 93.9 & 73.7 & 84.6 & 39.5 & 51.7 & 53.0 & 78.0 \\
OpenThinker2 (7B) & 16.0 & 20.0 & 16.8 & 28.0 & 14.0 & 20.0 & 2.8 & 4.0 & 12.4 & 18.0 & 65.3 & 96.0 & 60.5 & 97.0 & 79.1 & 84.6 & 42.3 & 58.6 & 58.9 & 84.0 \\
Skywork-OR1-Pre. (7B) & 14.4 & 20.0 & 12.5 & 12.0 & 11.7 & 24.0 & 1.6 & 0.0 & 10.0 & 14.0 & 61.6 & 88.0 & 55.9 & 78.8 & 74.3 & 92.3 & 36.9 & 48.3 & 54.2 & 74.0 \\
Skywork-OR1-Math (7B) & 17.4 & 20.0 & 17.1 & 20.0 & 13.6 & 28.0 & 0.9 & 0.0 & 12.2 & 17.0 & 67.9 & 92.0 & 67.4 & 93.9 & 76.6 & 92.3 & 47.6 & 62.1 & \underline{63.0} & 84.0 \\
AceMath-RL (7B) & 19.4 & 32.0 & 19.3 & 32.0 & 14.4 & 24.0 & 3.5 & 4.0 & \underline{14.2} & \underline{23.0} & 69.7 & 96.0 & 63.7 & 93.9 & 79.0 & 84.6 & 44.2 & 69.0 & 61.5 & \underline{86.0} \\
OpenMath-Nemo. (7B) & 26.9 & 36.0 & 18.6 & 28.0 & 19.8 & 28.0 & 4.4 & 4.0 & \textbf{{17.4}} & \textbf{{24.0}} & 86.4 & 100.0 & 76.4 & 97.0 & 91.5 & 100.0 & 55.3 & 72.4 & \textbf{{74.7}} & \textbf{{91.0}} \\

\midrule

DS-R1-Distill (14B) & 16.1 & 16.0 & 17.0 & 16.0 & 18.1 & 32.0 & 2.1 & 4.0 & 13.3 & 17.0 & 69.0 & 96.0 & 65.1 & 97.0 & 79.4 & 92.3 & 44.0 & 65.5 & 61.8 & \underline{87.0} \\
Light-R1-DS (14B) & 21.8 & 24.0 & 22.2 & 28.0 & 17.8 & 36.0 & 2.6 & 4.0 & \underline{16.1} & \underline{23.0} & 72.3 & 88.0 & 73.0 & 100.0 & 84.3 & 92.3 & 47.6 & 65.5 & \underline{66.9} & 86.0 \\
OpenMath-Nemo. (14B) & 28.7 & 40.0 & 22.1 & 32.0 & 21.0 & 32.0 & 3.4 & 4.0 & \textbf{{18.8}} & \textbf{{27.0}} & 87.9 & 100.0 & 78.5 & 93.9 & 95.8 & 100.0 & 59.9 & 86.2 & \textbf{{77.7}} & \textbf{{94.0}} \\

\midrule

Qwen3 (30B-A3B, Think) & 38.8 & 44.0 & 33.8 & 44.0 & 26.7 & 36.0 & 5.9 & 4.0 & \textbf{26.3} & \textbf{32.0} & 91.4 & 100.0 & 92.9 & 100.0 & 90.9 & 92.3 & 75.6 & 93.1 & \textbf{87.2} & \textbf{97.0} \\

\midrule

DS-R1-Distill (32B) & 22.4 & 32.0 & 21.4 & 24.0 & 20.3 & 40.0 & 3.4 & 4.0 & 16.9 & 25.0 & 73.6 & 100.0 & 71.8 & 97.0 & 84.5 & 92.3 & 49.0 & 69.0 & 67.3 & 89.0 \\
QwQ (32B) & 32.9 & 28.0 & 26.6 & 36.0 & 26.7 & 44.0 & 6.2 & 4.0 & \underline{23.1} & 28.0 & 91.8 & 100.0 & 87.0 & 100.0 & 95.0 & 100.0 & 69.0 & 89.7 & \textbf{84.0} & \textbf{97.0} \\
Light-R1-DS (32B) & 28.9 & 44.0 & 31.1 & 52.0 & 24.1 & 36.0 & 5.2 & 8.0 & 22.3 & \underline{35.0} & 84.2 & 100.0 & 83.3 & 100.0 & 92.5 & 100.0 & 62.1 & 82.8 & 78.6 & \underline{95.0} \\
OpenThinker2 (32B) & 24.1 & 32.0 & 22.9 & 32.0 & 18.0 & 20.0 & 2.6 & 4.0 & 16.9 & 22.0 & 79.4 & 96.0 & 74.0 & 100.0 & 90.4 & 92.3 & 56.5 & 79.3 & 72.4 & 92.0 \\
Skywork-OR1-Pre. (32B) & 37.2 & 52.0 & 32.3 & 48.0 & 27.0 & 40.0 & 4.2 & 4.0 & \textbf{25.2} & \textbf{36.0} & 89.3 & 100.0 & 87.3 & 100.0 & 92.4 & 100.0 & 63.9 & 82.8 & \underline{81.7} & \underline{95.0} \\
\textcolor{gray} {GLM-Z1-Air (32B)} & 35.0 & 44.0 & 21.5 & 32.0 & 19.5 & 24.0 & 4.5 & 4.0 & 20.1 & 26.0 & 86.5 & 100.0 & 79.5 & 90.9 & 90.4 & 100.0 & 59.1 & 75.9 & 76.8 & 90.0 \\
\textcolor{gray} {OpenMath-Nemo. (32B)} & 22.0 & 36.0 & 21.0 & 28.0 & 20.0 & 24.0 & 3.5 & 4.0 & 16.6 & 23.0 & 75.5 & 100.0 & 60.6 & 90.9 & 89.4 & 100.0 & 42.2 & 69.0 & 62.7 & 88.0 \\

\midrule

\textcolor{gray} {Qwen3 (235B-A22B, Think)} & 48.0 & 52.0 & 49.5 & 60.0 & 38.0 & 36.0 & 10.5 & 16.0 & \textbf{36.5} & \textbf{41.0} & 93.5 & 100.0 & 92.4 & 100.0 & 99.0 & 100.0 & 81.9 & 93.1 & \textbf{90.5} & \textbf{98.0} \\

\midrule

\textcolor{gray} {DeepSeek R1} & 30.0 & 40.0 & 25.5 & 32.0 & 18.5 & 24.0 & 4.0 & 4.0 & 19.5 & 25.0 & 90.5 & 100.0 & 82.2 & 97.0 & 94.2 & 100.0 & 60.8 & 72.4 & 79.6 & 91.0 \\
\textcolor{gray} {OpenAI o3-mini (high)} & 29.5 & 32.0 & 29.0 & 44.0 & 49.5 & 60.0 & 17.0 & 20.0 & \underline{31.2} & \underline{39.0} & 93.0 & 92.0 & 89.8 & 100.0 & 97.1 & 100.0 & 89.2 & 96.6 & \underline{91.4} & \textbf{97.0} \\
\textcolor{gray} {Gemini 2.5 Pro Exp 0325} & 71.5 & 76.0 & 75.5 & 84.0 & 59.0 & 72.0 & 27.5 & 36.0 & \textbf{{58.4}} & \textbf{{67.0}} & 92.0 & 100.0 & 97.0 & 100.0 & 98.1 & 100.0 & 84.5 & 89.7 & \textbf{{92.2}} & \textbf{{97.0}} \\
\bottomrule
\end{tabular}
}

\caption{{All models' performance on OlymMATH (EN). Models within each model size group are sorted by release time. The abbreviations ``Alg.'', ``Geo.'', ``Num.'', and ``Com.'' represent the four categories in OlymMATH. Highest accuracy per model size is bolded. The second highest accuracy per model size is underlined. Models sampled only 8 times are marked in \textcolor{gray}{gray} to indicate potential instability.}}
\label{performance-en-full}
\end{table*}

\begin{table*}
\centering
\setlength{\tabcolsep}{1.5pt} 

\label{tab:olym-zh}
\resizebox{\linewidth}{!}{
\begin{tabular}{l c >{\columncolor{consbg}}c c >{\columncolor{consbg}}c c >{\columncolor{consbg}}c c >{\columncolor{consbg}}c c >{\columncolor{consbg}}c | c >{\columncolor{consbg}}c c >{\columncolor{consbg}}c c >{\columncolor{consbg}}c c >{\columncolor{consbg}}c c >{\columncolor{consbg}}c }
\toprule
\multirow{3.5}{*}{\textbf{Model}} & \multicolumn{10}{c|}{\textbf{OlymMATH-HARD (ZH)}} & \multicolumn{10}{c}{\textbf{OlymMATH-EASY (ZH)}} \\
\cmidrule(lr){2-11} \cmidrule(lr){12-21}
& \multicolumn{2}{c}{\textbf{Alg.}} & \multicolumn{2}{c}{\textbf{Geo.}} & \multicolumn{2}{c}{\textbf{Num.}} & \multicolumn{2}{c}{\textbf{Com.}} & \multicolumn{2}{c|}{\textbf{Avg.}} & \multicolumn{2}{c}{\textbf{Alg.}} & \multicolumn{2}{c}{\textbf{Geo.}} & \multicolumn{2}{c}{\textbf{Num.}} & \multicolumn{2}{c}{\textbf{Com.}} & \multicolumn{2}{c}{\textbf{Avg.}} \\
& \texttt{P@1} & \texttt{C@k} & \texttt{P@1} & \texttt{C@k} & \texttt{P@1} & \texttt{C@k} & \texttt{P@1} & \texttt{C@k} & \texttt{P@1} & \texttt{C@k} & \texttt{P@1} & \texttt{C@k} & \texttt{P@1} & \texttt{C@k} & \texttt{P@1} & \texttt{C@k} & \texttt{P@1} & \texttt{C@k} & \texttt{P@1} & \texttt{C@k} \\
\midrule
Qwen3 (0.6B, Think) & 2.6 & 4.0 & 0.8 & 0.0 & 4.4 & 4.0 & 0.0 & 0.0 & \textbf{1.9} & \textbf{2.0} & 9.9 & 8.0 & 2.8 & 3.0 & 12.0 & 15.4 & 1.3 & 3.4 & \textbf{5.4} & \textbf{6.0} \\

\midrule

DS-R1-Distill (1.5B) & 1.8 & 0.0 & 1.3 & 0.0 & 1.1 & 0.0 & 0.0 & 0.0 & 1.0 & 0.0 & 13.7 & 20.0 & 6.3 & 9.1 & 20.9 & 30.8 & 2.6 & 0.0 & 9.0 & 12.0 \\
STILL-3-Pre. (1.5B) & 2.9 & 0.0 & 2.2 & 0.0 & 4.5 & 4.0 & 0.2 & 0.0 & 2.5 & 1.0 & 15.9 & 32.0 & 7.4 & 18.2 & 27.6 & 46.2 & 4.3 & 6.9 & 11.3 & \underline{22.0} \\
DeepScaleR-Pre. (1.5B) & 4.4 & 8.0 & 2.6 & 4.0 & 6.4 & 8.0 & 0.1 & 0.0 & \underline{3.4} & \underline{5.0} & 15.9 & 20.0 & 7.2 & 9.1 & 32.6 & 46.2 & 8.9 & 20.7 & \underline{13.2} & 20.0 \\
OpenMath-Nemo. (1.5B) & 13.9 & 16.0 & 9.8 & 4.0 & 13.3 & 16.0 & 0.8 & 0.0 & \textbf{9.5} & \textbf{9.0} & 67.9 & 96.0 & 37.6 & 57.6 & 65.3 & 76.9 & 27.6 & 41.4 & \textbf{45.9} & \textbf{65.0} \\

\midrule

Qwen3 (4B, Think) & 12.5 & 20.0 & 7.0 & 8.0 & 12.6 & 24.0 & 0.9 & 0.0 & \textbf{8.3} & \textbf{13.0} & 70.8 & 88.0 & 61.0 & 75.8 & 74.8 & 92.3 & 41.8 & 51.7 & \textbf{59.7} & \textbf{74.0} \\

\midrule

DS-R1-Distill (7B) & 6.1 & 8.0 & 7.9 & 12.0 & 6.6 & 8.0 & 0.6 & 0.0 & 5.3 & \underline{7.0} & 38.0 & 64.0 & 30.8 & 51.5 & 49.2 & 61.5 & 18.7 & 27.6 & 31.5 & 49.0 \\
Light-R1-DS (7B) & 7.1 & 4.0 & 9.4 & 12.0 & 7.8 & 12.0 & 1.1 & 0.0 & 6.3 & \underline{7.0} & 42.9 & 76.0 & 42.7 & 72.7 & 56.9 & 61.5 & 22.7 & 31.0 & 38.8 & 60.0 \\
OpenThinker2 (7B) & 7.0 & 0.0 & 7.3 & 8.0 & 7.4 & 8.0 & 1.0 & 0.0 & 5.7 & 4.0 & 48.2 & 80.0 & 44.7 & 72.7 & 57.8 & 76.9 & 22.4 & 37.9 & 40.8 & 65.0 \\
Skywork-OR1-Pre. (7B) & 4.7 & 4.0 & 7.8 & 8.0 & 7.4 & 8.0 & 0.4 & 0.0 & 5.1 & 5.0 & 41.1 & 60.0 & 36.6 & 54.5 & 58.1 & 69.2 & 23.6 & 34.5 & 36.8 & 52.0 \\
Skywork-OR1-Math (7B) & 6.4 & 8.0 & 8.3 & 8.0 & 9.8 & 12.0 & 0.8 & 0.0 & 6.3 & \underline{7.0} & 45.2 & 72.0 & 40.0 & 63.6 & 62.3 & 69.2 & 30.2 & 37.9 & 41.3 & 59.0 \\
AceMath-RL (7B) & 6.4 & 8.0 & 10.7 & 12.0 & 7.8 & 8.0 & 1.4 & 0.0 & \underline{6.6} & \underline{7.0} & 55.1 & 88.0 & 46.6 & 75.8 & 66.9 & 76.9 & 31.0 & 44.8 & \underline{46.9} & \underline{70.0} \\
OpenMath-Nemo. (7B) & 25.0 & 32.0 & 20.8 & 28.0 & 22.3 & 36.0 & 4.8 & 4.0 & \textbf{18.2} & \textbf{25.0} & 86.8 & 100.0 & 72.7 & 90.9 & 91.8 & 100.0 & 57.9 & 79.3 & \textbf{74.4} & \textbf{91.0} \\

\midrule

DS-R1-Distill (14B) & 5.2 & 0.0 & 5.3 & 4.0 & 8.7 & 16.0 & 0.2 & 0.0 & 4.9 & 5.0 & 43.1 & 56.0 & 38.9 & 66.7 & 58.2 & 69.2 & 24.8 & 31.0 & 38.4 & 54.0 \\
Light-R1-DS (14B) & 6.2 & 4.0 & 7.5 & 8.0 & 10.9 & 12.0 & 0.2 & 0.0 & \underline{6.2} & \underline{6.0} & 56.6 & 84.0 & 45.5 & 75.8 & 66.5 & 76.9 & 28.7 & 37.9 & \underline{46.1} & \underline{67.0} \\
OpenMath-Nemo. (14B) & 28.7 & 32.0 & 26.1 & 40.0 & 26.8 & 40.0 & 4.2 & 4.0 & \textbf{21.4} & \textbf{29.0} & 88.3 & 100.0 & 75.2 & 100.0 & 94.5 & 100.0 & 60.2 & 86.2 & \textbf{76.6} & \textbf{96.0} \\

\midrule

Qwen3 (30B-A3B, Think) & 35.6 & 40.0 & 24.1 & 28.0 & 18.1 & 24.0 & 2.7 & 4.0 & \textbf{20.1} & \textbf{24.0} & 87.8 & 92.0 & 84.7 & 97.0 & 91.3 & 100.0 & 61.9 & 65.5 & \textbf{79.7} & \textbf{87.0} \\

\midrule

DS-R1-Distill (32B) & 6.5 & 0.0 & 5.4 & 4.0 & 10.6 & 12.0 & 0.7 & 0.0 & 5.8 & 4.0 & 45.2 & 52.0 & 41.8 & 63.6 & 60.2 & 69.2 & 26.0 & 37.9 & 40.4 & 54.0 \\
QwQ (32B) & 20.9 & 24.0 & 15.9 & 16.0 & 17.6 & 24.0 & 2.0 & 0.0 & 14.1 & 16.0 & 85.4 & 96.0 & 76.6 & 97.0 & 92.9 & 100.0 & 53.8 & 69.0 & \textbf{74.3} & \textbf{89.0} \\
Light-R1-DS (32B) & 16.8 & 28.0 & 12.0 & 12.0 & 13.4 & 16.0 & 4.4 & 16.0 & 11.6 & \underline{18.0} & 70.1 & 96.0 & 64.1 & 93.9 & 80.4 & 92.3 & 39.8 & 51.7 & 60.7 & 82.0 \\
OpenThinker2 (32B) & 13.6 & 16.0 & 11.1 & 16.0 & 12.7 & 20.0 & 0.9 & 0.0 & 9.6 & 13.0 & 68.0 & 92.0 & 64.3 & 93.9 & 84.6 & 92.3 & 44.8 & 65.5 & 62.2 & 85.0 \\
Skywork-OR1-Pre. (32B) & 19.6 & 20.0 & 16.8 & 20.0 & 18.9 & 24.0 & 3.5 & 4.0 & \underline{14.7}  & 17.0 & 79.5 & 96.0 & 72.1 & 93.9 & 88.0 & 100.0 & 45.4 & 58.6 & \underline{68.3} & 85.0 \\
\textcolor{gray}{GLM-Z1-Air (32B)} & 18.0 & 16.0 & 12.0 & 8.0 & 16.0 & 16.0 & 2.5 & 4.0 & 12.1 & 11.0 & 76.0 & 96.0 & 69.3 & 78.8 & 89.4 & 92.3 & 41.8 & 48.3 & 65.6 & 76.0 \\
\textcolor{gray}{OpenMath-Nemo. (32B)} & 22.5 & 36.0 & 22.5 & 32.0 & 22.5 & 28.0 & 3.5 & 4.0 & \textbf{17.8} & \textbf{25.0} & 68.0 & 96.0 & 62.5 & 90.9 & 90.4 & 100.0 & 48.7 & 72.4 & 63.5 & \underline{88.0} \\

\midrule

\textcolor{gray}{Qwen3 (235B-A22B, Think)} & 36.5 & 48.0 & 43.5 & 48.0 & 28.5 & 32.0 & 4.0 & 8.0 & \textbf{28.1} & \textbf{34.0} & 91.0 & 100.0 & 90.2 & 97.0 & 94.2 & 100.0 & 78.4 & 89.7 & \textbf{87.5} & \textbf{96.0} \\

\midrule

\textcolor{gray}{DeepSeek R1} & 20.0 & 24.0 & 25.0 & 28.0 & 17.0 & 16.0 & 1.5 & 0.0 & 15.9 & 17.0 & 79.5 & 96.0 & 74.6 & 84.8 & 88.5 & 92.3 & 49.6 & 55.2 & 70.4 & 80.0 \\
\textcolor{gray}{OpenAI o3-mini (high)} & 31.5 & 40.0 & 32.5 & 44.0 & 48.5 & 56.0 & 19.0 & 28.0 & \underline{32.9} & \underline{42.0} & 93.0 & 96.0 & 89.4 & 100.0 & 99.0 & 100.0 & 85.8 & 93.1 & \underline{90.5} & \textbf{97.0} \\
\textcolor{gray}{Gemini 2.5 Pro Exp 0325} & 65.0 & 76.0 & 78.0 & 80.0 & 53.5 & 56.0 & 25.0 & 40.0 & \textbf{\underline{55.4}} & \textbf{\underline{63.0}} & 90.5 & 96.0 & 93.2 & 93.9 & 100.0 & 100.0 & 84.1 & 86.2 & \textbf{\underline{90.8}} & {\underline{93.0}} \\
\bottomrule
\end{tabular}
}

\caption{All models' performance on OlymMATH (ZH). Models within each model size group are sorted by release time. The abbreviations ``Alg.'', ``Geo.'', ``Num.'', and ``Com.'' represent the four categories in OlymMATH. Highest accuracy per model size is bolded. The second highest accuracy per model size is underlined. Models sampled only 8 times are marked in \textcolor{gray}{gray} to indicate potential instability.  }
\label{performance-zh-full}
\end{table*}

\begin{table*}
\centering
\small

\begin{tblr}{
  width = \linewidth, 
  colspec = {Q[c,m,wd=0.16\linewidth] X[c,m] X[c,m]}, 
  cell{1}{2} = {halign=c}, 
  cell{1}{3} = {halign=c}, 
  cell{2}{1} = {r=2}{c},
  cell{4}{1} = {r=4}{c}, 
  hline{1,Z} = {0.08em, solid}, 
  hline{2}   = {0.05em, solid}, 
  hline{4}   = {0.05em, solid}  
}
     & \textbf{Type} & \textbf{Example} \\ 
Included & {Real number} &$2^{2017}+\arctan 2$ \\
         & {Interval} & $[\sqrt{33}, +\infty)$, $(4, 5\pi]$ \\
Excluded & {Set Operations} & $\{4, 5\}\cup \{1, 8\}$ \\
         & {Variable} & $\sqrt[3]{5}a^2$, $p^2-pq$ \\
         & {Complex number} & $9+4\mathrm{i}$, $\sqrt{-4}$ \\
         & {Text} & $\text{East}$, $\text{Alice}$ \\
\end{tblr}
\caption{\emph{Included} and \emph{excluded} formats of the answer.}
\label{format}
\end{table*}